\documentclass[11pt]{article}

\usepackage[final]{acl}

\usepackage{times}
\usepackage{latexsym}

\usepackage[T1]{fontenc}

\usepackage[utf8]{inputenc}

\usepackage{microtype}

\usepackage{inconsolata}

\usepackage{graphicx}

\usepackage[table]{xcolor}

\newcommand{\eg}{\textit{e}.\textit{g}.}

\usepackage{amsmath}
\usepackage{mathtools}
\usepackage{amsthm}
\usepackage{amssymb}

\usepackage{tabularx}
\usepackage{booktabs} 
\usepackage{multirow}

\usepackage{arydshln}
\usepackage[linesnumbered,ruled,vlined]{algorithm2e} 
\usepackage{tcolorbox}
\usepackage{array}
\usepackage{makecell}
\usepackage{bm}
\usepackage{graphicx}
\usepackage{marvosym}
\usepackage{footmisc}

\usepackage{caption}
\usepackage{float}

\title{GARL: Game-Theoretic Reinforcement Learning for Multi-Agent Strategic Prioritisation}

\author{
 \textbf{Yuxiao Ye\textsuperscript{1}\thanks{Equal contribution.}},
 \textbf{Yiwen Zhang\textsuperscript{2}\footnotemark[1]\thanks{Work done during an internship at THUNLP.}},
 \textbf{Huiyuan Xie\textsuperscript{1}},
\\
 \textbf{Yuqin Huang\textsuperscript{1}},
 \textbf{Zhiyuan Liu\textsuperscript{1}\thanks{Corresponding author.}},
\\
\\
 \textsuperscript{1}Tsinghua University
\\
 \textsuperscript{2}The Hong Kong University of Science and Technology (Guangzhou)
\\
 \small{
   \{yeyuxiao, liuzy\}@mail.tsinghua.edu.cn, yzhang452@connect.hkust-gz.edu.cn
 }
}

\begin{document}
\maketitle

\begin{abstract}

LLM-based multi-agent systems are increasingly used for strategic decision-making tasks.
In such settings, performance depends not only on individual model capabilities, but also on the policies by which agents interact and adapt.
Multi-agent reinforcement learning can optimise these interaction policies, but its reward design often remains task-specific and weakly grounded in interaction structure.
To address this gap, we propose \textbf{GARL}, a \textbf{GA}me-theoretic \textbf{R}einforcement \textbf{L}earning framework for multi-agent strategic prioritisation.
GARL formalises strategic prioritisation as a two-stage game: competing agents first allocate strategic resources over a shared candidate set, and a higher-level arbiter then produces the final ranking.
The resulting game-theoretic utilities are converted into role-specific reinforcement signals, allowing policy optimisation to be guided by structured interaction.
We instantiate GARL on issues-in-dispute ranking, where the goal is to prioritise core issues in legal proceedings.
Experiments show that GARL improves ranking performance, enables small open-source LLMs to become competitive with a strong closed-source LLM under the same candidate-ranking setting, and yields gains in legal-domain competence and broader strategic decision-making.
Overall, GARL demonstrates how game-theoretic interaction structure can be turned into reinforcement-learning objectives, providing a principled approach to policy optimisation in multi-agent strategic prioritisation.
\end{abstract}

\section{Introduction}

Large language model (LLM)-based multi-agent systems (MASs) have become an important paradigm for complex tasks. 
Their performance depends not only on individual model capability but also on how agents strategically interact and coordinate \citep{dorri2018multi,guo2024large,ye2026linguagame}.
In this work, we focus on a practically important class of problems in multi-agent strategic interaction: \emph{multi-agent strategic prioritisation}.
In such problems, multiple agents hold different preferences over a shared candidate set, and the goal is to determine which candidates should receive higher priority for downstream decision-making \citep{macarthur2011multi,turner2018distributed,creech2021resource}.
This problem appears in many settings where agents compete or negotiate over what should matter most before a final decision is made, such as legal proceedings, public policy prioritisation, and investment allocation.

Existing work on LLM-based MASs has studied interaction modelling for strategic decision-making, but often focuses on workflow orchestration or inference-time coordination rather than direct policy optimisation \citep{Chen2024agentverse,qian-etal-2024-chatdev,wu2024autogen}.
Multi-agent reinforcement learning (MARL) directly optimises agent policies, yet its reward design often remains task-specific and weakly grounded in interaction structure \citep{albrecht2024multi,liu2026llm,wan2026rema}.
This leaves a gap for multi-agent strategic prioritisation: interaction structure can provide a principled basis for organising role-specific learning signals, but is rarely used to guide policy optimisation in MARL.

To address this gap, we propose \textbf{GARL}, a \textbf{GA}me-theoretic \textbf{R}einforcement \textbf{L}earning framework for multi-agent strategic prioritisation.
GARL formalises strategic prioritisation as a two-stage game.
In the \emph{Agenda Allocation} stage, agents allocate strategic resources over a shared candidate set to shape candidate salience.
In the \emph{Arbitration} stage, a higher-level arbiter produces the final ranking conditioned on the induced strategic state.
Instead, it organises reinforcement signals through the game-theoretic structure of the interaction itself, tying rewards to the allocation and arbitration utilities induced by the two-stage game.
This also reduces reliance on annotated supervision targets for learning strategic interaction policies.

We instantiate GARL on the legal task of issues-in-dispute ranking.
Issues in dispute are contested legal or factual questions, such as whether a contract is valid, whether a breach occurred, or how damages should be calculated.
In legal proceedings, different parties may foreground different issues, while the judge must determine which ones are central and adjudicatively meaningful enough to structure the proceedings \citep{menkel2004legal}.
In our instantiation, the prosecution and the defence allocate argumentative resources over candidate issues, and the judge produces the final ranking.
This setting provides a concrete instantiation of multi-agent strategic prioritisation, combining a shared candidate set, competing strategic roles, and a constrained final arbiter.

We evaluate this instantiation of GARL through a multi-level empirical design.
At the core-task level, we test whether GARL improves issues-in-dispute ranking.
Beyond the target task, we further examine broader legal-domain competence and strategic decision-making capability.
Across these evaluations, GARL consistently improves ranking performance, enables smaller open-source LLMs to become competitive with a strong closed-source LLM under the same candidate-ranking setting, and provides supporting evidence of gains beyond the target task.

Our contributions are as follows:

\begin{itemize}
    \item We propose GARL, a game-theoretic reinforcement learning framework for multi-agent strategic prioritisation. 

    \item We introduce a two-stage game formulation of allocation and arbitration for multi-agent strategic prioritisation, and show how the resulting utilities can organise rewards in MARL, reducing reliance on annotated supervision for strategic interaction learning.

    \item We instantiate GARL on issues-in-dispute ranking and demonstrate strong gains on the target ranking task. We further provide supporting evidence of gains in broader legal-domain competence and broader strategic decision-making.
\end{itemize}

\section{Related Work}

We consider our work related to three lines of research: LLM-based MAS, game-theoretic interaction modelling, and MARL and self-play training.

\paragraph{LLM-based MAS.}

Recent LLM-based multi-agent systems have shown that complex tasks can benefit from structured agent interaction \citep{chen2024agentcourt,Chen2024agentverse,cui2026chatlaw}.
For example, ChatDev \citep{qian-etal-2024-chatdev} frames software development as chat-based collaboration among role-specialised agents, while MetaGPT \citep{hong2024metagpt} incorporates software-engineering standard operating procedures into agent collaboration through specialised roles and structured intermediate artifacts.

These systems organise labour division, information exchange, and workflow handoff, but do not explicitly model strategic interaction among agents.
By contrast, our work focuses on strategic interaction in MASs, where agents may have distinct objectives and interaction is not merely a communication mechanism, but part of the decision structure itself.

\paragraph{Game-Theoretic Interaction Modelling.}

Another related line of work models LLM interaction through game-theoretic perspectives, including strategic gameplay, negotiation, and communication \citep{peters2024contingency,he2025generative}.
For example, Cicero \citep{meta2022human} combines controllable dialogue generation with strategic reasoning for Diplomacy.
\citet{hua2024game} further use classical game-theoretic concepts (\eg \  Nash-equilibrium reasoning) to guide LLM agents in games.

Most relevant to our work is LinguaGame \citep{ye2026linguagame}, which models multi-agent dialogue generation as a signalling game over communicative intents and strategies.
It uses training-free equilibrium approximation for inference-time utterance selection.
GARL shares the intuition that game-theoretic structure can organise multi-agent interaction, but shifts the focus from communication-level inference-time adjustment to RL-based policy optimisation through role-specific game utilities.

\begin{figure*}[t]
\centering
\includegraphics[width=0.98\textwidth]{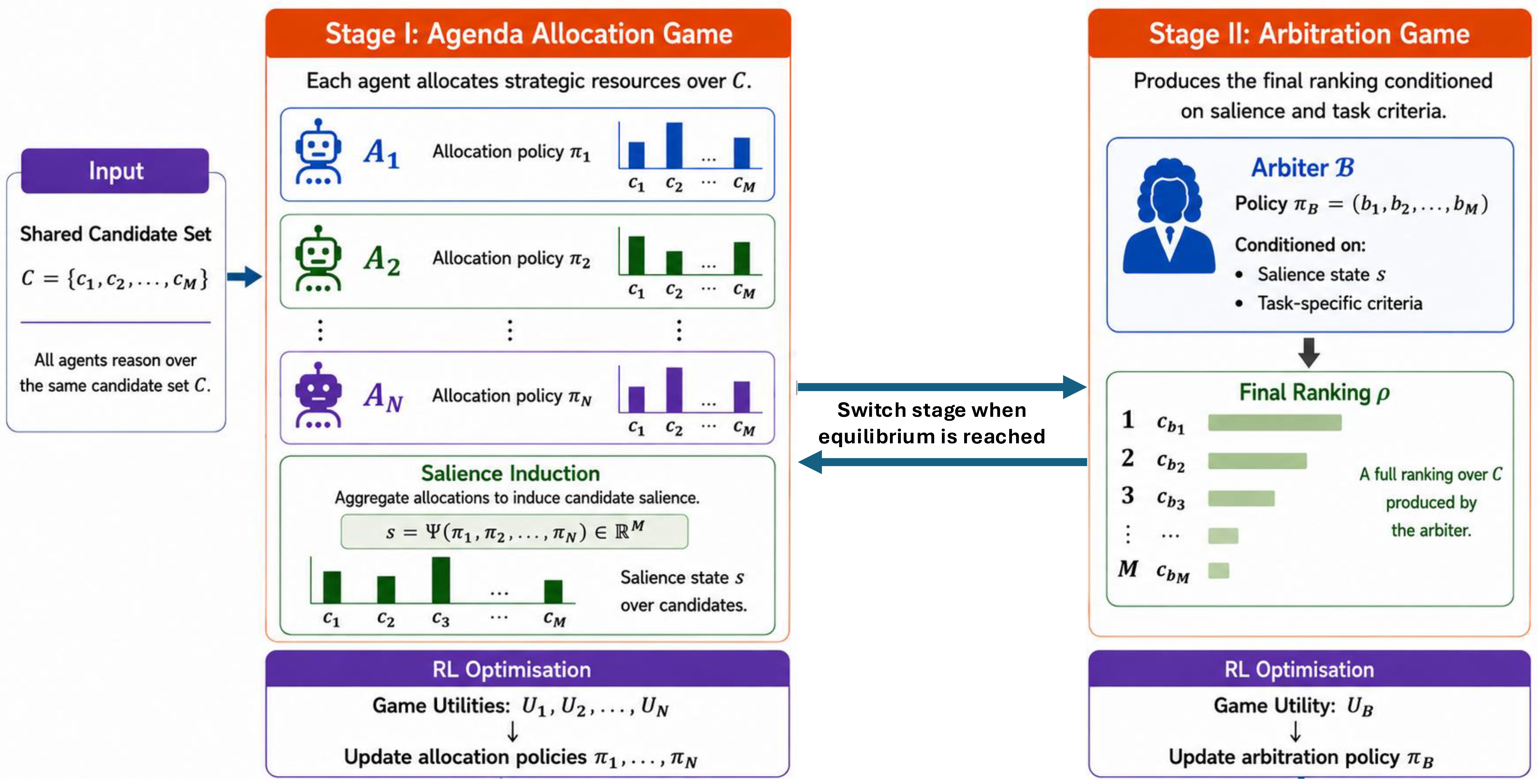}
\caption{
Overview of GARL. 
GARL alternates between agenda allocation, where allocator agents induce candidate salience, and arbitration, where an arbiter agent ranks candidates conditioned on salience and task-specific criteria.
Utilities from both stages provide role-specific reinforcement signals for policy optimisation.
}
\label{figs/framework}
\end{figure*} 

\paragraph{MARL and Self-Play Training.}

Recent work has begun to extend RL for LLMs from single-agent settings to multi-turn, multi-agent interaction \citep{park2025maporl,sarkar2025training}.
A particularly relevant direction uses self-play in strategic environments to train stronger models \citep{chen2024self,zhang2024survey}.
For example, SPAG~\citep{cheng2024self} and SPIRAL~\citep{liu2025spiral} both train LLMs through self-play in adversarial or zero-sum game environments, using game outcomes or online opponent improvement to induce transferable reasoning capabilities.
MARSHAL \citep{yuan2025marshal} extends this line by introducing turn-level advantage estimation and agent-specific normalisation to stabilise multi-agent RL and improve generalisable reasoning.

These works are closely related to ours in connecting strategic multi-agent interaction with RL.
However, their primary emphasis is capability acquisition through self-play, where strategic games serve as training environments.
GARL differs in its modelling target: it formalises the target task family itself as a game, with utilities that define role-specific MARL rewards.

\section{Method}

We propose GARL, studying multi-agent strategic prioritisation through a two-stage game formulation and its RL-based optimisation.

\subsection{Problem Formulation and Framework Overview}

We formalise multi-agent strategic prioritisation as a ranking problem over a shared candidate set.
Let \(\mathcal{C}=\{c_1,\ldots,c_M\}\) denote the candidate set, where each \(c_i\) is a possible object of prioritisation.
A set of agents interact over \(\mathcal{C}\), potentially preferring different candidates to receive greater attention, higher priority, or final selection.
The output is a ranking \(\rho\in\Pi(\mathcal{C})\), where \(\Pi(\mathcal{C})\) denotes the set of all permutations over \(\mathcal{C}\).

GARL specifies an allocation--arbitration skeleton for this process, as illustrated in Figure~\ref{figs/framework}.
In many strategic prioritisation settings, final ranking is preceded by a stage in which agents compete to shape what becomes salient.
GARL therefore decomposes prioritisation into a two-stage game: (1) an \textbf{Agenda Allocation Game}, in which agents allocate strategic resources over the shared candidate set to shape candidate salience; and (2) an \textbf{Arbitration Game}, in which a higher-level arbiter produces the final ranking conditioned on the induced salience state.
The abstract game utilities are defined according to this allocation--arbitration structure, with concrete semantics supplied by the target task.
They are used as role-specific reinforcement signals, enabling utility-driven policy optimisation without final-ranking supervision.

\subsection{Agenda Allocation Game}

The first stage of GARL is the \emph{Agenda Allocation Game}.
It is inspired by \emph{agenda setting games} in game theory \citep{zhu1992issue}, in which agents strategically shape which candidates become more salient.
In our setting, this strategic process operates over a fixed candidate set.

Let \(\mathcal{A}=\{A_1,\ldots,A_N\}\) be the set of allocating agents.
Each agent \(A_j\) chooses an allocation policy \(\pi_j=(a_{j1},\ldots,a_{jM})\), where \(a_{ji}\) denotes the proportion of strategic resources allocated by \(A_j\) to candidate \(c_i\).

Given all allocation policies, the environment induces a shared salience distribution $s=\Psi(\pi_1,\ldots,\pi_N)$, where \(s_i\) denotes the allocation-induced salience of \(c_i\).
The salience function \(\Psi(\cdot)\) should capture strategic interdependence, so that candidate salience is jointly shaped by competing allocations rather than passively revealed from a single agent's preference.
Each allocating agent then receives a utility \(U_j\) that depends on the induced salience state and the arbiter's current policy.
This produces the intermediate strategic state used by the subsequent Arbitration Game.

In our instantiation for issues-in-dispute ranking, the candidate set is \(\mathcal{I}=\{f_1,\ldots,f_M\}\), where each \(f_i\) is a candidate issue in dispute.
The allocating agents are the prosecution \(P\) and the defence \(D\), with allocation policies \(\pi_P=(a_{P1},\ldots,a_{PM})\) and \(\pi_D=(a_{D1},\ldots,a_{DM})\).
To instantiate \(\Psi(\cdot)\), each candidate issue first receives an unnormalised contention score:
\begin{equation}
\hat{s}_i
=
(a_{Pi}+a_{Di})
\left(
1-\frac{|a_{Pi}-a_{Di}|}{a_{Pi}+a_{Di}+\epsilon}
\right),
\end{equation}
where \(\epsilon>0\) is a small stabilisation constant.
The first term measures total argumentative attention, while the second rewards balanced engagement between the two parties.
The final salience distribution is obtained by normalisation: $s_i=\frac{\hat{s}_i}{\sum_j \hat{s}_j}$.

The party-side utility is defined as
\begin{equation}
U_X
=
\frac{1}{M}\sum_{i=1}^{M} v_{Xi}(b_i+s_i),
\end{equation}
where \(X\in\{P,D\}\), \(b_i\) is the judge's current prioritisation score for \(f_i\), and \(v_{Xi}\) is the party-specific strategic value of \(f_i\).
Specifically, \(v_{Xi}\) estimates the party-side advantage of arguing the case primarily around issue \(f_i\).
It is pre-computed before training through LLM-based issue-level win-probability estimation, and is then used as additional inputs to the allocation game.
The corresponding prompt is provided in Figure~\ref{figs/pm_win} in Appendix~\ref{app:prompts}.
In this way, each party optimises both agenda salience and expected judicial recognition.

\subsection{Arbitration Game}

The second stage is the \emph{Arbitration Game}, where a higher-level arbiter produces the final ranking conditioned on the strategic state induced by allocation.
This creates a \emph{Stackelberg-style} dependency \citep{fang2021introduction} between agents' strategic allocation and the final decision process.

Let \(B\) denote the arbiter.
Given \(\mathcal{C}\), salience distribution \(s\), and task-specific criteria \(\kappa\), the arbiter produces a ranking policy $\pi_B(\mathcal{C}; s; \kappa)=(b_{1},\ldots,b_{M})$, where \(b_i\) is the prioritisation score assigned to candidate \(c_i\).
The arbiter then receives a utility \(U_B\) evaluating how well this ranking serves the target task while remaining conditioned on the strategic state generated by competing agents.

In our instantiation, the arbiter is the judge \(J\), and the task-specific criteria \(Q\) are defined as judicial adequacy criteria derived through consultation with practising judges and lawyers.
These criteria include relevance, definiteness, provability, materiality, and legality (see Figure~\ref{figs/pm_qi} for details).
The judge's prioritisation policy is then defined as \(\pi_J(\mathcal{I};s;Q) = (b_1,\ldots,b_M)\), where \(b_i\) denotes the prioritisation score assigned to issue \(f_i\).
Let \(q_i\) denote the judicial adequacy of \(f_i\) under \(Q\), obtained from an external LLM-as-a-judge scorer.
The judge utility is then instantiated as:
\begin{equation}
U_J
=
\frac{1}{M}\sum_{i=1}^{M}
\Bigl(
b_i q_i
-
|b_i-\max(a_{Pi},a_{Di})|
\Bigr).
\end{equation}
The first term rewards higher scores for legally adequate issues, while the second penalises deviations from issues receiving strong strategic emphasis from either party.
Thus, the judge's objective is grounded in both judicial adequacy and the strategic emphasis induced by the parties' allocations.

\subsection{Reinforcement Learning Optimisation} \label{sub:reinforcement_learning_optimisation}

GARL turns the role-specific utilities induced by the allocation--arbitration game into reinforcement signals.
This connects the game structure to RL-based policy optimisation.
The following optimisation procedure is described under our issues-in-dispute ranking instantiation, while remaining applicable to other GARL instantiations.

\paragraph{Alternating policy optimisation.}
In our instantiation, the prosecution and the defence share an allocator model \(\pi_{\theta_{PD}}\) optimised via \(U_P\) and \(U_D\), while the judge uses an arbiter model \(\pi_{\theta_J}\) optimised via \(U_J\).
RL training alternates between these two models within each case batch.
Each model is updated until a KL-based equilibrium condition or a maximum number of update steps is reached, after which training switches to the other model.
Because \(P\) and \(D\) share the allocator parameters \(\theta_{PD}\), their role-specific gradients are accumulated before a single optimiser step.
The arbiter model is updated independently.
Algorithm~\ref{GARL} summarises the full process.

\begin{algorithm}[!t]
\small
\caption{GARL Training with Equilibrium-Based Switching}
\label{GARL}
\KwIn{
  case batches \(\mathcal{D}\),
  shared allocator model \(\pi_{\theta_{PD}}\),
  arbiter model \(\pi_{\theta_J}\),
  total steps \(N\),
  maximum inner steps \(K\),
  thresholds \(\tau_{PD},\tau_J\).
}
\KwOut{Updated parameters \(\theta_{PD},\theta_J\).}

\(step \leftarrow 0\); \(phase \leftarrow PD\); \(k \leftarrow 0\)\;
\(\mathrm{new\_batch} \leftarrow \mathrm{True}\)\;
\(\mathrm{eq}_{PD} \leftarrow \mathrm{False}\); \(\mathrm{eq}_J \leftarrow \mathrm{False}\)\;

\While{\(step < N\)}{
  \If{\(\mathrm{new\_batch}\)}{
    sample case batch \(d\sim\mathcal{D}\)\;
    \(\mathrm{eq}_{PD} \leftarrow \mathrm{False}\); \(\mathrm{eq}_J \leftarrow \mathrm{False}\)\;
    \(\mathrm{new\_batch} \leftarrow \mathrm{False}\)\;
  }

  \tcp{Rollout}
  generate \(\pi_P,\pi_D\sim\pi_{\theta_{PD}}(\cdot)\)\;
  compute salience distribution \(s=\Psi(\pi_P,\pi_D)\)\;
  generate judge policy \(\pi^J\sim\pi_{\theta_J}(\cdot)\)\;
  compute \(U_P,U_D,U_J\)\;
  convert role utilities to advantages\;
  compute \(\mathrm{KL}_P,\mathrm{KL}_D,\mathrm{KL}_J\)\;

  \tcp{Update}
  \uIf{\(phase=PD\)}{
    \If{\(k<K \land \neg\mathrm{eq}_{PD}\)}{
      compute \(\nabla_P\) and \(\nabla_D\)\;
      accumulate \(
      \nabla_{PD}
      =
      \omega_P\nabla_P
      +
      \omega_D\nabla_D
      \)\;
      update \(\theta_{PD}\) once using \(\nabla_{PD}\)\;
      \(\mathrm{eq}_{PD} \leftarrow (\mathrm{KL}_P\le\tau_{PD})\land(\mathrm{KL}_D\le\tau_{PD})\)\;
      \(k\leftarrow k+1\); \(step\leftarrow step+1\)\;
    }
    \Else{
      \(phase\leftarrow J\); \(k\leftarrow 0\)\;
    }
  }
  \ElseIf{\(phase=J\)}{
    \If{\(k<K \land \neg\mathrm{eq}_J\)}{
      compute \(\nabla_{\theta_J}\)\;
      update \(\theta_J\)\;
      \(\mathrm{eq}_J \leftarrow \mathrm{KL}_J\le\tau_J\)\;
      \(k\leftarrow k+1\); \(step\leftarrow step+1\)\;
    }
    \Else{
      \(phase\leftarrow PD\); \(k\leftarrow 0\)\;
      \(\mathrm{new\_batch}\leftarrow \mathrm{True}\)\;
    }
  }
}
\end{algorithm}

\paragraph{From turn-level utilities to token-level advantages.}
The role utilities are defined at the turn level, whereas our policy-gradient optimisation requires token-level learning signals.
We therefore assign each role utility to the last valid token of the corresponding role response, thereby forming a sparse token-level reward for policy optimisation.
The resulting sparse rewards are converted into token-level advantages in Algorithm~\ref{GARL},  following REINFORCE++~\cite{hu2025reinforce++,hu2025reinforce}.

Detailed formulations of the GARL objective, KL-based equilibrium metric, and token-level advantage computation are provided in Appendix~\ref{sec:additional_details_of_reinforcement_learning_optimisation}.

\section{Evaluation Design} \label{sec:evaluation_design}

We evaluate our issues-in-dispute ranking instantiation of GARL through a three-level design.
Across all three levels, we compare base models with their GARL-trained counterparts to examine both task-specific gains and broader capability effects.

\paragraph{Level I: Issues-in-dispute ranking.}
Level I evaluates the target task of issues-in-dispute ranking.
For each case, the system is given a fixed candidate pool \(\mathcal{I}\), with gold issues forming an unordered subset \(G\subseteq \mathcal{I}\).
The model outputs a complete ranking \(\rho\) over all candidates.

We report two metrics.
First, Recall@\(k\) with \(k=|G|\) measures set-level recovery under a gold-size selection budget:
\[
\mathrm{Recall@}k
=
\frac{|G\cap \mathrm{Top}_k(\rho)|}{|G|},
\qquad k=|G|.
\]
Since \(k=|G|\), Precision@\(k\) and F1@\(k\) are equivalent to Recall@\(k\), so we report only Recall@\(k\).

Second, mean average precision (mAP) measures full-ranking quality.
For each case, average precision is defined as
\[
\mathrm{AP}(\rho,G)
=
\frac{1}{|G|}
\sum_{j=1}^{M}
\frac{|G\cap \mathrm{Top}_j(\rho)|}{j}
\cdot
\mathbf{1}[\rho_j \in G],
\]
where \(\rho_j\) denotes the candidate ranked at position \(j\) in \(\rho\).
Then mAP is obtained by averaging AP over all test cases.

\paragraph{Level II: Legal-domain competence.}
Level II evaluates broader legal-domain competence beyond the target task using LawBench \citep{fei2024lawbench}.
LawBench covers 20 Chinese legal tasks organised into three cognitive levels: legal knowledge memorisation, legal knowledge understanding, and legal knowledge application.

\paragraph{Level III: Broader strategic decision-making.}
Level III evaluates broader strategic decision-making capability beyond the legal domain using GameBench \citep{costarelli2024gamebench}.
GameBench evaluates LLM agents across multiple game environments involving strategic reasoning under interaction, including hidden information, language communication, social deduction, cooperation, and non-deterministic outcomes.

\section{Experiments}

We conduct experiments to evaluate GARL's instantiation for issues-in-dispute ranking, following the three-level evaluation design.

\subsection{Data}

We use a pre-release version of LexIssue \citep{xie2026lexissue}, a contemporaneous benchmark for issues-in-dispute ranking, to construct the data for GARL training and Level I evaluation.
This version contains 10 causes of action and over 600 cases.
Each LexIssue instance pairs case information with a set of gold issues. 
The case information includes party information, claims, factual statements, defence responses, and evidence summaries. 
The gold issues are treated as an unordered set, with around three gold issues per case on average.

For training, we first randomly select five causes of action and remove cases with only one gold issue, resulting in 263 original training cases.
For each remaining case, we construct three candidate pools: one with only gold issues, one adding an equal number of LLM-generated distractors, and one replacing all gold issues with the same number of LLM-generated issues.
The generated candidates are produced by Qwen3.5-122B~\citep{qwen3.5}, using the prompt in Figure~\ref{figs/pm_gen}.
The three variants are mixed and randomly shuffled, yielding \(263\times3\) training instances.
Gold annotations are used only for candidate-pool construction; models do not observe which candidates are gold issues during GARL training.

For Level I testing, we construct both in-domain and out-of-domain splits.
The in-domain split uses the same 263 cases as training, while the out-of-domain split contains 350 cases sampled from causes of action disjoint from the five training causes.
This design allows us to evaluate both optimisation on cases encountered during RL training and transfer to unseen cases and legal categories.
For each test case, we retain all gold issues and add Qwen3.5-122B-generated distractors until the candidate pool contains exactly 10 issues.
Although the in-domain cases overlap with the training cases, their test candidate pools differ, and no supervised ranking labels are used during training.

\begin{table*}[t]
    \scriptsize
    \renewcommand{\arraystretch}{1}
    \centering
    \begin{tabular}{>{\centering\arraybackslash}p{0.07\linewidth}>{\centering\arraybackslash}p{0.03\linewidth}>{\centering\arraybackslash}p{0.04\linewidth}>{\centering\arraybackslash}p{0.05\linewidth}>{\centering\arraybackslash}p{0.07\linewidth}>{\centering\arraybackslash}p{0.065\linewidth}>{\centering\arraybackslash}p{0.05\linewidth}>{\centering\arraybackslash}p{0.06\linewidth}>{\centering\arraybackslash}p{0.06\linewidth}>{\centering\arraybackslash}p{0.05\linewidth}>{\centering\arraybackslash}p{0.06\linewidth}>{\centering\arraybackslash}p{0.06\linewidth}}\toprule
         &     &GPT-4& DeepSeek-R1&Qwen-3.5-27B& Qwen2.5-7B-Instruct&Qwen3-4B&  Qwen3-4B-\(PD\) &  Qwen3-4B-\(J\)  & Qwen3-8B& Qwen3-8B-\(PD\) & Qwen3-8B-\(J\) \\\midrule
         \multirow{2}{*}{in/prompt}&     recall&80.92& 78.82&77.75& 39.16&65.95&  \cellcolor{yellow!20}68.94&   \cellcolor{yellow!20}68.65 & 76.94& \cellcolor{yellow!60}\textbf{83.14}& \cellcolor{yellow!20}81.46\\
 &  mAP&89.18& 87.52& 87.19& 44.01& 73.92& \cellcolor{yellow!20}78.86& \cellcolor{yellow!20}77.40 & 85.22& \cellcolor{yellow!60}\textbf{90.43}& \cellcolor{yellow!20}88.85\\
         \multirow{2}{*}{in/one-shot}&     recall&\textbf{80.07}& 66.86&73.11& 37.97&67.98&  \cellcolor{yellow!20}70.75&    65.79& 68.99& \cellcolor{yellow!20}74.40& \cellcolor{yellow!20}73.40\\
 &  mAP&\textbf{89.14}& 78.87& 84.59& 42.23& 77.57& \cellcolor{yellow!20}80.78&  75.96& 80.55& \cellcolor{yellow!20}84.35& \cellcolor{yellow!20}82.54\\
 
 \hdashline
 
         \multirow{2}{*}{out/prompt}&     recall&77.80& 76.46&72.56& 38.20&66.65&  \cellcolor{yellow!20}70.15&    \cellcolor{yellow!20}71.26& 74.78& \cellcolor{yellow!60}\textbf{78.46}& \cellcolor{yellow!20}76.26\\
 &  mAP&87.12& 85.94& 84.42& 43.63& 76.87& \cellcolor{yellow!20}80.63&  \cellcolor{yellow!20}80.35& 84.02& \cellcolor{yellow!60}\textbf{87.63}& \cellcolor{yellow!20}86.47\\ 
 \multirow{2}{*}{out/one-shot}&  recall&\textbf{76.12}& 65.46& 69.75& 36.36& 68.51& \cellcolor{yellow!20}70.45&  67.21& 70.55& 70.23& \cellcolor{yellow!20}70.68\\ 
 &  mAP&\textbf{86.29}& 77.63& 81.16& 40.50& 78.73& \cellcolor{yellow!20}80.73&  77.73& 81.20& \cellcolor{yellow!20}81.61& \cellcolor{yellow!20}81.73\\ \bottomrule
    \end{tabular}
    \caption{Main results on LexIssue for issues-in-dispute ranking.}
    \label{tab:lexissue_results}
\end{table*}

\begin{table*}[t]
\centering
\renewcommand{\arraystretch}{1.1}
\scriptsize
\begin{tabular}{>{\centering\arraybackslash}p{0.035\linewidth}>{\centering\arraybackslash}p{0.05\linewidth}>{\centering\arraybackslash}p{0.06\linewidth}>{\centering\arraybackslash}p{0.06\linewidth}>{\centering\arraybackslash}p{0.06\linewidth}>{\centering\arraybackslash}p{0.07\linewidth}>{\centering\arraybackslash}p{0.07\linewidth}>{\centering\arraybackslash}p{0.07\linewidth}>{\centering\arraybackslash}p{0.07\linewidth}>{\centering\arraybackslash}p{0.07\linewidth}>{\centering\arraybackslash}p{0.07\linewidth}}\toprule
 & GPT-4 & Qwen-Chat-7B& Fuzi-Mingcha& ChatLaw-33B& Qwen3-4B&  Qwen3-4B-\(PD\) &  Qwen3-4B-\(J\)  & Qwen3-8B& Qwen3-8B-\(PD\) & Qwen3-8B-\(J\)\\\midrule
\textbf{Overall} & 53.85& 38.99& 28.78& 29.14& 47.08& 46.94& \textbf{48.28}& 47.22& 46.92&\textbf{47.99}\\
\bottomrule
\end{tabular}
\caption{Overall results on LawBench.}
\label{tab:lawbench_results_all}
\end{table*}

\subsection{Models and Baselines}

We train GARL from Qwen3-4B and Qwen3-8B \citep{qwen3technicalreport}.
For each base model, GARL produces a \(PD\) model (the allocator) and a \(J\) model (the arbiter).
The two trained models are evaluated separately.

For Level I, we compare against GPT-4 \citep{achiam2023gpt} and open-source baselines including DeepSeek-R1 \citep{deepseekai2025deepseekr1incentivizingreasoningcapability}, Qwen2.5-7B-Instruct \citep{qwen2.5}, and Qwen3.5-27B \citep{qwen3.5}.
For Level II, we evaluate on LawBench and include representative legal-domain baselines reported in the benchmark.
For Level III, we evaluate base and GARL-trained models on GameBench.

\subsection{Training and Inference Setup}

Training follows the alternating policy optimisation procedure described in Section~\ref{sub:reinforcement_learning_optimisation}.
All trained models use the same data mixture, hyperparameter setting, and fixed random seeds.
Detailed hyperparameters are provided in Appendix~\ref{app:params}.
Prompts given to \(P\)/\(D\) and \(J\) during training are provided in Figures~\ref{figs/pm_pa} and~\ref{figs/pm_j}.
We use an external LLM (Qwen3.5-122B) to pre-compute utility inputs, namely win-probability \(v_{Xi}\) and judicial adequacy score \(q_i\).
This avoids directly exposing the trainable agents to domain-specific scoring criteria in prompts, enabling a cleaner evaluation of GARL.

At inference time for Level I, all models are evaluated under the same candidate-ranking format.
Each model is given the same ranking prompt (see Figure~\ref{figs/pm_test}) and directly outputs a ranking over the candidate issues.
We evaluate both direct prompting and one-shot prompting, with the one-shot example drawn from outside both the training and test sets.

\subsection{Evaluation Protocol}

For Level I, we report Recall@\(|G|\) and mAP for issues-in-dispute ranking on LexIssue.
For Level II, we use LawBench standard metrics and report overall and task-level results.
For Level III, we use GameBench standard metrics and report overall and per-game results, evaluating 4B and 8B models only within their respective size groups.

\section{Results and Discussion}

We organise the results around the three-level evaluation design: target-task performance, broader legal-domain competence, and broader strategic decision-making capability.
We then provide an additional analysis of training dynamics to inspect the strategic interaction learned during GARL training.

\paragraph{Issues-in-Dispute Ranking.}

Table~\ref{tab:lexissue_results} shows that GARL consistently improves the target ranking task, especially under direct prompting.
For Qwen3-4B, GARL with the \(PD\) model significantly\footnote{Significance in this paper is tested through two-tailed paired t-tests at $\alpha=0.01$.} improves mAP on both in-domain and out-of-domain splits.
The gains are larger for Qwen3-8B: \(PD\) significantly improves in-domain mAP from 85.22 to 90.43, surpassing GPT-4 under the same setting, and also improves out-of-domain mAP to a level comparable to GPT-4.
The out-of-domain gains suggest that GARL learns transferable prioritisation rather than merely adapting to seen cases.
Overall, \textbf{GARL substantially improves issues-in-dispute ranking and enables smaller open-source models, especially the 8B model, to approach or locally surpass a strong closed-source LLM}.
This indicates that GARL is effective for modelling and optimising issues-in-dispute ranking as a multi-agent strategic prioritisation problem.

The one-shot setting shows less stable gains, and in some cases even hurts the performance of base models compared with direct prompting. 
This suggests that the test-time example may introduce an in-context ranking bias rather than a consistently helpful demonstration. 
Since GARL is trained without demonstrations, this bias can interfere with the learned prioritisation policy. 
By contrast, direct prompting better exposes the learned policy, under which GARL yields more consistent improvements.

A role-dependent pattern also emerges: \(PD\) improves more consistently than \(J\).
We assume this is because LexIssue annotations are based primarily on pleadings from both parties rather than on full judicial reasoning. 
The gold issues therefore are primarily grounded in the dispute structure framed by the parties, while judicial perspective plays a more secondary role. 
This makes them more closely aligned with the party-side strategic salience learned by \(PD\). Although \(J\) also incorporates party-induced salience, its judicial-adequacy objective may make it less directly aligned with this party-oriented annotation setting.

\paragraph{Broader Legal-Domain Competence.}

Table~\ref{tab:lawbench_results_all} reports the overall LawBench results.
The \(J\) model consistently outperforms the corresponding base model, suggesting that arbiter-side training brings modest transfer to broader legal-domain competence.
For Qwen3-4B, the average score increases from 47.08 to 48.28 with \(J\), while \(PD\) remains comparable at 46.94.
For Qwen3-8B, \(J\) improves the average score from 47.22 to 47.99, while \(PD\) stays close to the base model at 46.92.
Thus, \textbf{GARL does not substantially degrade broader legal competence, and the arbiter model yields consistent overall gains across both model sizes}.

The task-level results in Appendix~\ref{app:results} show a role-dependent transfer pattern.
The \(J\) model improves more reliably on structured legal understanding and application tasks, consistent with its objective of legally adequate prioritisation.
By contrast, \(PD\) shows selective gains on content understanding, summarisation, and argument mining, but slightly declines overall.
This likely reflects objective mismatch: party-side allocation training emphasises strategic salience shaping, whereas many LawBench tasks require neutral answer selection, structured extraction, or format-compliant generation.

\paragraph{Broader Strategic Decision-Making Capability.}

Table~\ref{tab:gamebench_overall} reports the overall GameBench results, showing that \textbf{GARL improves overall strategic decision-making performance in both model size groups}.
For Qwen3-4B, GARL improves the overall rating from \(-0.08\) to 0.01 with the \(PD\) model and 0.07 with the \(J\) model, and improves the overall score from 0.47 to 0.52 and 0.53 respectively.
For Qwen3-8B, the \(PD\) model achieves the strongest gain, improving the overall rating from \(-0.14\) to 0.18 and the overall score from 0.45 to 0.57; the \(J\) model also improves the overall score to 0.52.

Detailed per-game results are provided in Table~\ref{tab:gamebench_detail} (in Appendix~\ref{app:results}).
They suggest role-dependent transfer effects: \(PD\) models tend to benefit games involving active priority shaping or resource allocation, while \(J\) models show gains in settings that require conditional judgement or signal integration.

\begin{table}[t]
\centering
\small
\renewcommand{\arraystretch}{1.1}
\begin{tabular}{lccc}
\toprule
 & \textbf{Base} & \textbf{PD} & \textbf{J} \\
\midrule
\multicolumn{4}{c}{\cellcolor{gray!10}\textbf{Qwen3-4B}} \\
Overall Rating & -0.08 & \underline{0.01} & \textbf{0.07} \\
Overall Score  & 0.47  & \underline{0.52} & \textbf{0.53} \\
\midrule
\multicolumn{4}{c}{\cellcolor{gray!10}\textbf{Qwen3-8B}} \\
Overall Rating & -0.14 & \textbf{0.18} & \underline{-0.04} \\
Overall Score  & 0.45  & \textbf{0.57} & \underline{0.52} \\
\bottomrule
\end{tabular}
\caption{Overall results on GameBench. }
\label{tab:gamebench_overall}
\end{table}

\paragraph{Training Dynamics}

Beyond benchmark performance, we further inspect GARL's training dynamics from two perspectives.
Figure~\ref{figs/delta_reward} in Appendix~\ref{app:td} reports block-level reward-change trajectories for \(P\), \(D\), and \(J\), while Figure~\ref{figs/PD_ratio} reports the step-level \(P\)--\(D\) loss ratio.

For reward-change analysis, we group training steps into blocks, where each block contains all update steps performed for the same case within a stage.
For a role \(r\in\{P,D,J\}\), the reward change of block \(m\) is defined as $\Delta R_r^{(m)}
=
R_{r,\mathrm{last}}^{(m)}
-
R_{r,\mathrm{first}}^{(m)}$, where \(R_{r,\mathrm{first}}^{(m)}\) and \(R_{r,\mathrm{last}}^{(m)}\) are the role rewards at the first and last update steps of the block.
Figure~\ref{figs/delta_reward} shows that these reward changes gradually shrink toward small fluctuations around zero, suggesting that case-specific optimisation blocks become locally stable rather than producing unbounded reward escalation.

Figure~\ref{figs/PD_ratio} further characterises the step-level coupling between the two party-side agents.
At each update step, we compute $r_{PD}
=
\mathrm{Avg}(\mathrm{Loss}_P) / \mathrm{Avg}(\mathrm{Loss}_D)$, where the loss sign is determined by the negative advantage.
Values below zero indicate opposite update directions, while values near \(-1\) indicate opposite directions with comparable strength.
Across training, \(r_{PD}\) gradually concentrates around \(-1\), suggesting balanced adversarial coupling between \(P\) and \(D\).

Together, these dynamics provide evidence of approximate strategic stabilisation and are consistent with equilibrium-seeking behaviour.

\section{Conclusion}

We presented GARL, a game-theoretic reinforcement learning framework for multi-agent strategic prioritisation.
GARL formalises prioritisation as an allocation--arbitration interaction: strategic agents allocate resources over candidate items to shape salience, and an arbiter produces the final prioritisation conditioned on the induced strategic state.
By deriving role-specific utilities from this interaction, GARL connects multi-agent game design with RL-based policy optimisation.

We instantiated GARL on issues-in-dispute ranking, where the prosecution and defence act as allocation agents and the judge acts as the arbiter.
Empirically, GARL substantially improves ranking performance, enabling smaller open-source models to approach or locally surpass a strong closed-source LLM under the same candidate-ranking setting.
Additional evaluations suggest that GARL can preserve or moderately improve broader legal-domain competence and induce broader strategic decision-making gains beyond the target task.

Overall, GARL shows that structured game-theoretic interaction can be operationalised as task-grounded reinforcement signals, offering a principled approach for optimising multi-agent strategic prioritisation.

\section*{Limitations and Future Directions}

\textbf{Scope of the framework.}
GARL is designed for multi-agent strategic prioritisation rather than all forms of multi-agent interaction.
It assumes a shared candidate set and a prioritisation objective, making it less directly applicable to open-ended collaboration, unconstrained dialogue, or general multi-agent planning.
A natural direction for future work is to relax the fixed-candidate assumption by allowing agents to propose, merge, or revise candidates during interaction.

\textbf{Task-specific utility design.}
GARL abstracts the allocation--arbitration interaction structure and optimisation protocol, but concrete utility components remain task-specific.
In our legal instantiation, strategic favourability and judicial adequacy must still be explicitly defined.
Future work may explore more reusable utility designs for recurring prioritisation settings, such as policy issue prioritisation, investment allocation, and other structured decision-making tasks.

\textbf{Scope of empirical validation.}
The present work provides a full instantiation of GARL on the task of ranking issues in dispute.
While the allocation--arbitration formulation can be naturally instantiated in principle in other strategic prioritisation settings, this paper empirically validates only the legal dispute-focus ranking scenario.
We provide an illustrative mapping between GARL components and several potential application scenarios in Table~\ref{tab:garl_scenario_mapping} in Appendix~\ref{app:scenario_mapping}.
Broader validation still requires applying GARL to more tasks and examining whether similar training dynamics and transfer effects hold across domains.

\textbf{Responsible AI considerations.}
This work studies multi-agent strategic prioritisation in legal settings as a research framework rather than as a deployable legal decision-making system.
Our instantiation on ranking issues in dispute is intended for controlled analysis of structured strategic interaction, not for replacing judicial reasoning or adjudicative authority in practice.
Because legal prioritisation is a high-stakes setting, any real-world deployment would require substantial additional validation, including careful assessment of fairness, bias, robustness, transparency, and alignment with legal and ethical standards.



\bibliography{custom}

@article{chen2024agentcourt,
  title={AgentCourt: Simulating Court with Adversarial Evolvable Lawyer Agents},
  author={Chen, Guhong and Fan, Liyang and Gong, Zihan and Xie, Nan and Li, Zixuan and Liu, Ziqiang and Li, Chengming and Qu, Qiang and Ni, Shiwen and Yang, Min},
  journal={CoRR},
  year={2024}
}

@inproceedings{Chen2024agentverse,
 author = {Chen, Weize and Su, Yusheng and Zuo, Jingwei and Yang, Cheng and Yuan, Chenfei and Chan, Chi-Min and Yu, Heyang and Lu, Yaxi and Hung, Yi-Hsin and Qian, Chen and Qin, Yujia and Cong, Xin and Xie, Ruobing and Liu, Zhiyuan and Sun, Maosong and Zhou, Jie},
 booktitle = {International Conference on Representation Learning},
 pages = {20094--20136},
 title = {AgentVerse: Facilitating Multi-Agent Collaboration and Exploring Emergent Behaviors},
 volume = {2024},
 year = {2024}
}

@article{cui2026chatlaw,
  title={Chatlaw: A Multi-Agent Legal Assistant based on a Role-Aligned Mixture-of-Experts Architecture},
  author={Cui, Jiaxi and Ning, Munan and Li, Zongjian and Li, Hao and Ya, Yang and Chen, Bohua and Ling, Bin and Tian, Yonghong and Yuan, Li},
  journal={Fundamental Research},
  year={2026},
  publisher={Elsevier}
}

@article{dorri2018multi,
  title={Multi-agent systems: A survey},
  author={Dorri, Ali and Kanhere, Salil S and Jurdak, Raja},
  journal={Ieee Access},
  volume={6},
  pages={28573--28593},
  year={2018},
  publisher={IEEE}
}

@inproceedings{guo2024large,
  title={Large Language Model Based Multi-agents: A Survey of Progress and Challenges},
  author={Guo, Taicheng and Chen, Xiuying and Wang, Yaqi and Chang, Ruidi and Pei, Shichao and Chawla, Nitesh V and Wiest, Olaf and Zhang, Xiangliang},
  booktitle={IJCAI},
  year={2024}
}

@article{ye2026linguagame,
  title={LinguaGame: A Linguistically Grounded Game-Theoretic Paradigm for Multi-Agent Dialogue Generation},
  author={Ye, Yuxiao and Zhang, Yiming and Ma, Yiran and Xie, Huiyuan and Zhu, Huining and Liu, Zhiyuan},
  journal={arXiv preprint arXiv:2601.04516},
  year={2026}
}

@phdthesis{macarthur2011multi,
  title={Multi-agent coordination for dynamic decentralised task allocation},
  author={Macarthur, Kathryn},
  year={2011},
  school={University of Southampton}
}

@phdthesis{turner2018distributed,
  title={Distributed task allocation optimisation techniques in multi-agent systems},
  author={Turner, Joanna},
  year={2018},
  school={Loughborough University}
}

@article{creech2021resource,
  title={Resource allocation in dynamic multiagent systems},
  author={Creech, Niall and Pacheco, Natalia Criado and Miles, Simon},
  journal={arXiv preprint arXiv:2102.08317},
  year={2021}
}

@inproceedings{qian-etal-2024-chatdev,
    title = "{C}hat{D}ev: Communicative Agents for Software Development",
    author = "Qian, Chen  and
      Liu, Wei  and
      Liu, Hongzhang  and
      Chen, Nuo  and
      Dang, Yufan  and
      Li, Jiahao  and
      Yang, Cheng  and
      Chen, Weize  and
      Su, Yusheng  and
      Cong, Xin  and
      Xu, Juyuan  and
      Li, Dahai  and
      Liu, Zhiyuan  and
      Sun, Maosong",
    booktitle = "Proceedings of the 62nd Annual Meeting of the Association for Computational Linguistics (Volume 1: Long Papers)",
    month = aug,
    year = "2024",
    publisher = "Association for Computational Linguistics",
    url = "https://aclanthology.org/2024.acl-long.810/",
    doi = "10.18653/v1/2024.acl-long.810",
    pages = "15174--15186"
}

@inproceedings{wu2024autogen,
  title={Autogen: Enabling next-gen LLM applications via multi-agent conversations},
  author={Wu, Qingyun and Bansal, Gagan and Zhang, Jieyu and Wu, Yiran and Li, Beibin and Zhu, Erkang and Jiang, Li and Zhang, Xiaoyun and Zhang, Shaokun and Liu, Jiale and others},
  booktitle={First conference on language modeling},
  year={2024}
}

@book{albrecht2024multi,
  title={Multi-agent reinforcement learning: Foundations and modern approaches},
  author={Albrecht, Stefano V and Christianos, Filippos and Sch{\"a}fer, Lukas},
  year={2024},
  publisher={MIT Press}
}

@inproceedings{liu2026llm,
  title={Llm collaboration with multi-agent reinforcement learning},
  author={Liu, Shuo and Liang, Zeyu and Lyu, Xueguang and Amato, Christopher},
  booktitle={Proceedings of the AAAI Conference on Artificial Intelligence},
  volume={40},
  number={38},
  pages={32150--32158},
  year={2026}
}

@article{wan2026rema,
  title={Rema: Learning to meta-think for llms with multi-agent reinforcement learning},
  author={Wan, Ziyu and Li, Yunxiang and Wen, Xiaoyu and Song, Yan and Wang, Hanjing and Yang, Linyi and Schmidt, Mark and Wang, Jun and Zhang, Weinan and Hu, Shuyue and others},
  journal={Advances in Neural Information Processing Systems},
  volume={38},
  pages={126621--126667},
  year={2026}
}

@article{menkel2004legal,
  title={From legal disputes to conflict resolution and human problem solving: Legal dispute resolution in a multidisciplinary context},
  author={Menkel-Meadow, Carrie},
  journal={Journal of Legal Education},
  volume={54},
  number={1},
  pages={7--29},
  year={2004},
  publisher={JSTOR}
}

@inproceedings{hong2024metagpt,
  title={MetaGPT: Meta programming for a multi-agent collaborative framework},
  author={Hong, Sirui and Zhuge, Mingchen and Chen, Jonathan and Zheng, Xiawu and Cheng, Yuheng and Wang, Jinlin and Zhang, Ceyao and Yau, Steven and Lin, Zijuan and Zhou, Liyang and others},
  booktitle={International Conference on Learning Representations},
  volume={2024},
  pages={23247--23275},
  year={2024}
}

@article{meta2022human,
  title={Human-level play in the game of diplomacy by combining language models with strategic reasoning},
  author={FAIR, Meta and Bakhtin, Anton and Brown, Noam and Dinan, Emily and Farina, Gabriele and Flaherty, Colin and Fried, Daniel and Goff, Andrew and Gray, Jonathan and Hu, Hengyuan and others},
  journal={Science},
  volume={378},
  number={6624},
  pages={1067--1074},
  year={2022},
  publisher={American Association for the Advancement of Science}
}

@article{hua2024game,
  title={Game-theoretic llm: Agent workflow for negotiation games},
  author={Hua, Wenyue and Liu, Ollie and Li, Lingyao and Amayuelas, Alfonso and Chen, Julie and Jiang, Lucas and Jin, Mingyu and Fan, Lizhou and Sun, Fei and Wang, William and others},
  journal={arXiv preprint arXiv:2411.05990},
  year={2024}
}

@article{cheng2024self,
  title={Self-playing adversarial language game enhances llm reasoning},
  author={Cheng, Pengyu and Dai, Yong and Hu, Tianhao and Xu, Han and Zhang, Zhisong and Han, Lei and Du, Nan and Li, Xiaolong},
  journal={Advances in Neural Information Processing Systems},
  volume={37},
  pages={126515--126543},
  year={2024}
}

@article{liu2025spiral,
  title={Spiral: Self-play on zero-sum games incentivizes reasoning via multi-agent multi-turn reinforcement learning},
  author={Liu, Bo and Guertler, Leon and Yu, Simon and Liu, Zichen and Qi, Penghui and Balcells, Daniel and Liu, Mickel and Tan, Cheston and Shi, Weiyan and Lin, Min and others},
  journal={arXiv preprint arXiv:2506.24119},
  year={2025}
}

@article{yuan2025marshal,
  title={Marshal: Incentivizing multi-agent reasoning via self-play with strategic llms},
  author={Yuan, Huining and Xu, Zelai and Tan, Zheyue and Yi, Xiangmin and Guang, Mo and Long, Kaiwen and Hui, Haojia and Li, Boxun and Chen, Xinlei and Zhao, Bo and others},
  journal={arXiv preprint arXiv:2510.15414},
  year={2025}
}

@article{peters2024contingency,
  title={Contingency games for multi-agent interaction},
  author={Peters, Lasse and Bajcsy, Andrea and Chiu, Chih-Yuan and Fridovich-Keil, David and Laine, Forrest and Ferranti, Laura and Alonso-Mora, Javier},
  journal={IEEE Robotics and Automation Letters},
  volume={9},
  number={3},
  pages={2208--2215},
  year={2024},
  publisher={IEEE}
}

@article{he2025generative,
  title={Generative ai for game theory-based mobile networking},
  author={He, Long and Sun, Geng and Niyato, Dusit and Du, Hongyang and Mei, Fang and Kang, Jiawen and Debbah, M{\'e}rouane and Han, Zhu},
  journal={IEEE Wireless Communications},
  volume={32},
  number={1},
  pages={122--130},
  year={2025},
  publisher={IEEE}
}

@inproceedings{park2025maporl,
  title={Maporl: Multi-agent post-co-training for collaborative large language models with reinforcement learning},
  author={Park, Chanwoo and Han, Seungju and Guo, Xingzhi and Ozdaglar, Asuman E and Zhang, Kaiqing and Kim, Joo-Kyung},
  booktitle={Proceedings of the 63rd Annual Meeting of the Association for Computational Linguistics (Volume 1: Long Papers)},
  pages={30215--30248},
  year={2025}
}

@article{sarkar2025training,
  title={Training language models for social deduction with multi-agent reinforcement learning},
  author={Sarkar, Bidipta and Xia, Warren and Liu, C Karen and Sadigh, Dorsa},
  journal={arXiv preprint arXiv:2502.06060},
  year={2025}
}

@article{chen2024self,
  title={Self-play fine-tuning converts weak language models to strong language models},
  author={Chen, Zixiang and Deng, Yihe and Yuan, Huizhuo and Ji, Kaixuan and Gu, Quanquan},
  journal={arXiv preprint arXiv:2401.01335},
  year={2024}
}

@article{zhang2024survey,
  title={A survey on self-play methods in reinforcement learning},
  author={Zhang, Ruize and Xu, Zelai and Ma, Chengdong and Yu, Chao and Tu, Wei-Wei and Tang, Wenhao and Huang, Shiyu and Ye, Deheng and Ding, Wenbo and Yang, Yaodong and others},
  journal={arXiv preprint arXiv:2408.01072},
  year={2024}
}

@article{hu2025reinforce++,
  title={Reinforce++: A simple and efficient approach for aligning large language models},
  author={Hu, Jian},
  journal={arXiv e-prints},
  pages={arXiv--2501},
  year={2025}
}

@article{hu2025reinforce,
  title={Reinforce++: Stabilizing critic-free policy optimization with global advantage normalization},
  author={Hu, Jian and Liu, Jason Klein and Xu, Haotian and Shen, Wei},
  journal={arXiv preprint arXiv:2501.03262},
  year={2025}
}

@article{zhu1992issue,
  title={Issue competition and attention distraction: A zero-sum theory of agenda-setting},
  author={Zhu, Jian-Hua},
  journal={Journalism quarterly},
  volume={69},
  number={4},
  pages={825--836},
  year={1992},
  publisher={Sage Publications Sage CA: Los Angeles, CA}
}

@article{fang2021introduction,
  title={Introduction to game theory},
  author={Fang, Fei and Liu, Shutian and Basak, Anjon and Zhu, Quanyan and Kiekintveld, Christopher D and Kamhoua, Charles A},
  journal={Game theory and machine learning for cyber security},
  pages={21--46},
  year={2021},
  publisher={Wiley Online Library}
}

@inproceedings{fei2024lawbench,
  title={Lawbench: Benchmarking legal knowledge of large language models},
  author={Fei, Zhiwei and Shen, Xiaoyu and Zhu, Dawei and Zhou, Fengzhe and Han, Zhuo and Huang, Alan and Zhang, Songyang and Chen, Kai and Yin, Zhixin and Shen, Zongwen and others},
  booktitle={Proceedings of the 2024 conference on empirical methods in natural language processing},
  pages={7933--7962},
  year={2024}
}

@article{costarelli2024gamebench,
  title={Gamebench: Evaluating strategic reasoning abilities of llm agents},
  author={Costarelli, Anthony and Allen, Mat and Hauksson, Roman and Sodunke, Grace and Hariharan, Suhas and Cheng, Carlson and Li, Wenjie and Clymer, Joshua and Yadav, Arjun},
  journal={arXiv preprint arXiv:2406.06613},
  year={2024}
}

@misc{qwen3.5,
    title  = {{Qwen3.5}: Towards Native Multimodal Agents},
    author = {{Qwen Team}},
    month  = {February},
    year   = {2026},
    url    = {https://qwen.ai/blog?id=qwen3.5}
}

@misc{qwen3technicalreport,
      title={Qwen3 Technical Report}, 
      author={Qwen Team},
      year={2025},
      eprint={2505.09388},
      archivePrefix={arXiv},
      primaryClass={cs.CL},
      url={https://arxiv.org/abs/2505.09388}, 
}

@article{achiam2023gpt,
  title={Gpt-4 technical report},
  author={Achiam, Josh and Adler, Steven and Agarwal, Sandhini and Ahmad, Lama and Akkaya, Ilge and Aleman, Florencia Leoni and Almeida, Diogo and Altenschmidt, Janko and Altman, Sam and Anadkat, Shyamal and others},
  journal={arXiv preprint arXiv:2303.08774},
  year={2023}
}

@misc{deepseekai2025deepseekr1incentivizingreasoningcapability,
      title={DeepSeek-R1: Incentivizing Reasoning Capability in LLMs via Reinforcement Learning}, 
      author={DeepSeek-AI},
      year={2025},
      eprint={2501.12948},
      archivePrefix={arXiv},
      primaryClass={cs.CL},
      url={https://arxiv.org/abs/2501.12948}, 
}

@misc{qwen2.5,
    title = {Qwen2.5: A Party of Foundation Models},
    url = {https://qwenlm.github.io/blog/qwen2.5/},
    author = {Qwen Team},
    month = {September},
    year = {2024}
}

@misc{xie2026lexissue,
  title = {LexIssue: Benchmarking Legal Issue Identification in Litigation},
  author = {Xie, Huiyuan and Huang, Yuqin and Hao, Zhicheng and Cai, Yida and Wang, Shaochun and Ye, Yuxiao and Liu, Zhenghao },
  year = {2026},
  note = {To be released in June, 2026}
}

\clearpage

\appendix

\section{Additional Details of Reinforcement Learning Optimisation}
\label{sec:additional_details_of_reinforcement_learning_optimisation}

\subsection{GARL Policy Optimisation Objective}

In the main text, Algorithm~\ref{GARL} describes the alternating optimisation procedure between the shared allocator model \(\pi_{\theta_{PD}}\) and the arbiter model \(\pi_{\theta_J}\).
Here we provide the concrete policy optimisation objective used in implementation.

The overall GARL objective combines the role-specific policy losses for the prosecution, the defence, and the judge, together with entropy regularisation and a KL penalty:
\begin{equation}
\small
\label{eq:garl-main-objective}
\begin{split}
\mathcal{L}_{\mathrm{GARL}}
=
\mathbb{E}
\bigg[
&
\sum_{X\in\{P,D\}}
L_{\mathrm{clip}}\bigl(\hat{A}_t^X,r_t^{PD}\bigr)
+
L_{\mathrm{clip}}\bigl(\hat{A}_t^J,r_t^{J}\bigr)
\\
&
-\lambda_{\mathrm{ent}}^{PD}H(\pi_{\theta_{PD}})
-\lambda_{\mathrm{ent}}^{J}H(\pi_{\theta_J})
\\
&
-\lambda_{\mathrm{KL}}^{PD}
D_{\mathrm{KL}}
\bigl(
\pi_{\theta_{PD}}^{\mathrm{old}}
\|
\pi_{\theta_{PD}}
\bigr)
\bigg],
\end{split}
\end{equation}
where \(L_{\mathrm{clip}}\) denotes the clipped policy-gradient loss, \(\hat{A}_t^X\) and \(\hat{A}_t^J\) are token-level advantages for party \(X\in\{P,D\}\) and the judge respectively, and \(r_t^{PD}\) and \(r_t^{J}\) denote the corresponding probability ratios used in the policy update.
The terms \(H(\pi_{\theta_{PD}})\) and \(H(\pi_{\theta_J})\) denote entropy regularisation for the allocator and arbiter models.
The KL penalty regularises the shared allocator model against its previous policy.

\subsection{KL-Based Equilibrium Metric}

The phase switching condition in Algorithm~\ref{GARL} is based on the KL divergence between consecutive policies.
For a batch with \(B\) sequences and maximum length \(L\), the batch-level KL metric is computed over valid tokens:
\begin{equation}
\small
\label{eq:kl-divergence}
\begin{split}
D_{\mathrm{KL}}
=
\frac{1}{B}
\sum_{i=1}^{B}
\left(
\frac{1}{\sum_{j=1}^{L}\mathbb{I}_{i,j}+\epsilon}
\sum_{j=1}^{L}
p_{i,j}^{\mathrm{old}}
\Delta \log p_{i,j}
\cdot
\mathbb{I}_{i,j}
\right),
\end{split}
\end{equation}
where
\[
\Delta \log p_{i,j}
=
\log(p_{i,j}^{\mathrm{old}}+\epsilon)
-
\log(\max(p_{i,j}^{\mathrm{new}},\epsilon)).
\]
Here, \(p_{i,j}^{\mathrm{old}}\) and \(p_{i,j}^{\mathrm{new}}\) denote the old and new token probabilities, \(\mathbb{I}_{i,j}\) masks valid non-padding positions, and \(\epsilon\) is a small numerical constant.
For the shared allocator model, equilibrium is reached when both party-side KL values fall below the threshold:
\[
\mathrm{eq}_{PD}
=
(\mathrm{KL}_P\le \tau_{PD})
\land
(\mathrm{KL}_D\le \tau_{PD}).
\]
For the arbiter model, equilibrium is reached when
\[
\mathrm{eq}_{J}
=
(\mathrm{KL}_J\le \tau_J).
\]

\subsection{Role-Batch Pipeline for Token-Level Advantage Computation}

Algorithm~\ref{batch} describes the pipeline used to convert turn-level role utilities into token-level advantages.
For each role \(r\in\{P,D,J\}\), the corresponding utility \(U_r\) is first converted into a turn-level reward, injected at the final valid token of the role response, and then processed into token-level advantages.

\begin{algorithm}[h]
\small
\caption{RoleBatchPipeline: Turn-Level Utility \(\rightarrow\) Token-Level Advantage}
\label{batch}
\KwIn{Role batch \(\mathcal{B}_r\), current policy \(\pi_\theta\), role \(r\in\{P,D,J\}\).}
\KwOut{Enriched role batch \(\mathcal{X}_r\).}

Compute \(G^{\mathrm{turn}}\) from turn-level rewards with \(\gamma_{\mathrm{turn}}\)\;
Initialize token-level scores \(S^{\mathrm{tok}}\leftarrow 0\)\;

\For{each sample \(b\)}{
  Find last valid token index \(j^\star\) where \(\mathrm{step\text{-}id}\neq -100\)\;
  Assign \(S^{\mathrm{tok}}_{b,j^\star}\leftarrow G^{\mathrm{turn}}_{b,0}\)\;
}

Set token-level rewards \(r^{\mathrm{tok}}\leftarrow S^{\mathrm{tok}}\)\;
\(M\leftarrow(\mathrm{step\text{-}id}\neq -100)\)\;

Compute REINFORCE++ returns by reverse recursion with mask reset:
\(G^{\mathrm{tok}}_{b,t}=r^{\mathrm{tok}}_{b,t}+\gamma_{\mathrm{tok}}G^{\mathrm{tok}}_{b,t+1}\), then
\(G^{\mathrm{tok}}_{b,t+1}\leftarrow G^{\mathrm{tok}}_{b,t+1}\cdot M_{b,t}\)\;

Compute advantages:
\(A=\mathrm{Whiten}(G^{\mathrm{tok}};M)\odot M\)\;

Compute old log-probabilities under current policy:
\(\log\pi_{\theta}^{\mathrm{old}}\)\;

Return
\[
\mathcal{X}_r
=
\mathcal{B}_r
\cup
\{
S^{\mathrm{tok}},
r^{\mathrm{tok}},
G^{\mathrm{tok}},
A,
\log\pi_{\theta}^{\mathrm{old}}
\}.
\]
\end{algorithm}

The pipeline contains three implementation steps.
First, turn reward injection assigns the turn-level reward exclusively to the last valid token of each role response, yielding a sparse token-level reward signal.
Second, REINFORCE++ mask reset multiplies the running discounted sum by a validity mask at each step, thereby resetting reward accumulation when a response terminates or padding begins.
Third, masked whitening computes the mean and standard deviation only over valid tokens, mitigating distortion from variable-length sequences and ensuring that padded positions contribute zero gradient.

\clearpage

\section{Training Hyperparameters}
\label{app:params}

\begin{center}
\begin{minipage}{\textwidth}
\centering
\small
\begin{tabular}{p{0.42\textwidth} p{0.50\textwidth}}
\toprule
\textbf{Parameter} & \textbf{Value} \\
\midrule
\texttt{data.train\_batch\_size} & 8 \\
\texttt{data.max\_\{prompt,response\}\_length} & 4096, 2048 \\
\texttt{actor.ppo\_\{mini,micro\}\_batch\_size} & 8, 1 \\
\texttt{ref/rollout.log\_prob\_micro\_bsz} & 1 \\
\texttt{rollout.max\_num\_batched\_tokens} & 5120 \\
\texttt{actor.optim.lr} & \(1\times10^{-6}\) \\
\texttt{algorithm.adv\_estimator} & \texttt{reinforce\_plus\_plus} \\
\texttt{algorithm.\{gamma,lam\}\_token\_level} & 1.0, 1.0 \\
\texttt{algorithm.gamma\_turn\_level} & 1.0 \\
\texttt{trainer.total\_training\_steps} & 600 \\
\texttt{trainer.max\_inner\_steps} & 5 \\
\texttt{rollout.max\_num\_turns} & 1 \\
\texttt{rollout.temperature} & 0.5 \\
\(\omega_{P}\), \(\omega_{D}\) & 0.5, 0.5 \\
\texttt{algorithm.equilibrium\_threshold} 
& Qwen3-4B: 0.0002; Qwen3-8B: \(PD\) 0.00008, \(J\) 0.0006 \\
\bottomrule
\end{tabular}
\captionof{table}{Hyperparameters in the GARL training pipeline.}
\label{tab:hyperparameters}
\end{minipage}
\end{center}

\clearpage

\section{Detailed Results for Level II and III}
\label{app:results}

\begin{center}
\begin{minipage}{\textwidth}
\centering
\renewcommand{\arraystretch}{1.1}
\scriptsize
\begin{tabular}{>{\centering\arraybackslash}p{0.035\linewidth}>{\centering\arraybackslash}p{0.05\linewidth}>{\centering\arraybackslash}p{0.06\linewidth}>{\centering\arraybackslash}p{0.06\linewidth}>{\centering\arraybackslash}p{0.06\linewidth}>{\centering\arraybackslash}p{0.07\linewidth}>{\centering\arraybackslash}p{0.07\linewidth}>{\centering\arraybackslash}p{0.07\linewidth}>{\centering\arraybackslash}p{0.07\linewidth}>{\centering\arraybackslash}p{0.07\linewidth}>{\centering\arraybackslash}p{0.07\linewidth}}
\toprule
& GPT-4 & Qwen-Chat-7B & Fuzi-Mingcha & ChatLaw-33B & Qwen3-4B & Qwen3-4B-\(PD\) & Qwen3-4B-\(J\) & Qwen3-8B & Qwen3-8B-\(PD\) & Qwen3-8B-\(J\)\\
\midrule
1-1& 17.21& 17.73& 20.21& 15.98& 19.38& \cellcolor{yellow!20}20.21& 17.93& 21.76& \cellcolor{yellow!20}22.33&17.91\\
1-2& 54.8& 28.6& 12.8& 29.4& 60.4& \cellcolor{yellow!20}61.4& \cellcolor{yellow!20}60.8& 62.8& 62.6&\cellcolor{yellow!20}65.0\\
\hdashline
2-1  &  18.31&  25.16&  2.86&  3.67&  15.84&  15.31&  \cellcolor{yellow!20}23.38 & 22.90& 22.0&\cellcolor{yellow!20}27.66\\
2-2  &  46.0&  27.4&  2.4&  8.04&  36.4&  35.3&  \cellcolor{yellow!20}38.13 & 38.83& \cellcolor{yellow!20}41.0&\cellcolor{yellow!20}39.87\\
2-3  &  69.99&  32.96&  17.44&  32.08&  36.54&  36.48&  \cellcolor{yellow!20}37.39 & 42.07& 40.11&\cellcolor{yellow!20}45.88\\
2-4  &  44.4&  31.2&  8.8&  19.80&  38.0&  36.80&  \cellcolor{yellow!20}39.40 & 40.40& \cellcolor{yellow!20}40.60&\cellcolor{yellow!20}41.60\\
2-5  &  64.8&  46.71&  93.35&  37.16&  76.35&  \cellcolor{yellow!20}76.89&  \cellcolor{yellow!20}77.80& 78.07& 77.14&75.28\\
2-6  &  79.96&  57.34&  42.28&  30.14&  43.42&  \cellcolor{yellow!20}47.83&  \cellcolor{yellow!20}51.67 & 38.30& \cellcolor{yellow!20}38.92&\cellcolor{yellow!20}38.43\\
2-7  &  40.52&  42.58&  31.43&  35.47&  40.46&  \cellcolor{yellow!20}40.63&  35.92 & 20.94& \cellcolor{yellow!20}21.58&\cellcolor{yellow!20}23.24\\
2-8  &  59.0&  26.8&  11.4&  26.40&  49.60&  \cellcolor{yellow!20}50.60&  47.40 & 51.80& 51.60&51.40\\
2-9  &  76.55&  50.63&  21.26&  22.14&  59.80{\tiny(8\%)}&  \cellcolor{yellow!20}58.94{\tiny(7.4\%)}&  \cellcolor{yellow!20}60.06{\tiny(4\%)} & 62.77(5\%)& 62.45(5.2\%)&\cellcolor{yellow!20}66.41(2.2\%)\\
2-10 &  65.26&  21.27&  7.04&  10.56&  29.01{\tiny(9\%)}&  \cellcolor{yellow!20}27.33{\tiny(8.8\%)}&  \cellcolor{yellow!20}30.95{\tiny(3.4\%)} & 32.55(5.6\%)& 30.59(6.4\%)&\cellcolor{yellow!20}32.64(4.2\%)\\
\hdashline
3-1& 53.2& 52.86& 3.86& 25.99& 70.45& 67.66& \cellcolor{yellow!20}71.38& 74.28& 73.78&73.17\\
3-2& 33.15& 34.49& 32.96& 33.96& 33.49& \cellcolor{yellow!20}34.53& 33.36& 33.57& 32.25&28.57\\
3-3& 41.3& 39.91& 43.6& 12.24& 44.86& 43.80& \cellcolor{yellow!20}51.10& 44.22& 43.23&41.70\\
3-4& 83.21& 78.47& 78.95& 74.31& 82.12& \cellcolor{yellow!20}82.15& 80.42& 82.93& \cellcolor{yellow!20}83.34&81.58\\
3-5& 82.74& 73.92& 79.0& 73.01& 74.99& \cellcolor{yellow!20}75.38& \cellcolor{yellow!20}75.95& 73.93& \cellcolor{yellow!20}75.08&\cellcolor{yellow!20}81.85\\
3-6& 49.6& 26.8& 13.8& 34.20& 50.0& 48.8&\cellcolor{yellow!20}50.2 & 55.40& 54.80&55.0\\
3-7& 77.0& 44.6& 38.2& 41.60& 65.60& 65.40&\cellcolor{yellow!20}66.80 & 54.60& 52.80&\cellcolor{yellow!20}57.0\\
3-8& 19.9& 20.39& 13.95& 16.55& 14.90& 13.43&\cellcolor{yellow!20}15.60 & 12.24& 12.19&\cellcolor{yellow!20}15.59\\ 
\midrule
\textbf{Overall} & 53.85& 38.99& 28.78& 29.14& 47.08& 46.94& \textbf{48.28}& 47.22& 46.92&\textbf{47.99}\\
\bottomrule
\end{tabular}
\captionof{table}{Detailed results on LawBench.}
\label{tab:lawbench_results_detail}
\end{minipage}
\end{center}

\vspace*{20pt}

\begin{center}
\begin{minipage}{\textwidth}
\centering
\renewcommand{\arraystretch}{1.1}
\scriptsize
\begin{tabular}{c>{\centering\arraybackslash}p{0.06\linewidth}>{\centering\arraybackslash}p{0.06\linewidth}>{\centering\arraybackslash}p{0.06\linewidth}>{\centering\arraybackslash}p{0.06\linewidth}>{\centering\arraybackslash}p{0.06\linewidth}>{\centering\arraybackslash}p{0.06\linewidth}>{\centering\arraybackslash}p{0.06\linewidth}>{\centering\arraybackslash}p{0.06\linewidth}>{\centering\arraybackslash}p{0.06\linewidth}>{\centering\arraybackslash}p{0.06\linewidth}}
\toprule
& \textbf{Overall}& ALS& ARC& AYT& CN& HV& PT& SN& TRB& SB\\
\midrule
\multicolumn{11}{c}{\cellcolor{gray!10}Rating}\\
Qwen3-4B& -0.08 & -0.15 & \underline{-0.07} & \underline{-0.09} & \underline{-0.07} & \underline{-1.45} & \underline{0.22} & \textbf{1.24} & \underline{0.03} & -0.07 \\
Qwen3-4B-PD& \underline{0.01} & \textbf{0.28} & -0.14 & -1.03 & \textbf{0.36} & \textbf{3.11} & -0.45 & -2.32 & -0.38 & \textbf{0.06} \\
Qwen3-4B-J& \textbf{0.07} & \underline{-0.13} & \textbf{0.21} & \textbf{1.12} & -0.29 & -1.67 & \textbf{0.23} & \underline{1.07} & \textbf{0.35} & \underline{0.01} \\
\hdashline
Qwen3-8B& -0.14 & \underline{-0.14} & \underline{-0.00} & \underline{-0.13} & \underline{-0.10} & \textbf{0.15} & \textbf{0.15} & -1.31 & -0.22 & \textbf{0.00} \\
Qwen3-8B-PD& \textbf{0.18} & -0.16 & \textbf{0.30} & \textbf{0.52} & \textbf{0.30} & \underline{-0.07} & -0.29 & \textbf{1.72} & \underline{-0.00} & \textbf{0.00} \\
Qwen3-8B-J& \underline{-0.04} & \textbf{0.30} & -0.29 & -0.40 & -0.20 & -0.08 & \underline{0.14} & \underline{-0.40} & \textbf{0.22} & \textbf{0.00} \\
\multicolumn{11}{c}{\cellcolor{gray!10}Score}\\
Qwen3-4B& 0.47 & 0.45 & 0.48 & \underline{0.47} & \underline{0.47} & 0.30 & \textbf{0.57} & \textbf{0.51} & \underline{0.51} & 0.47 \\
Qwen3-4B-PD& \underline{0.52} & \textbf{0.60} & \underline{0.48} & 0.30 & \textbf{0.60} & \textbf{0.95} & 0.35 & 0.47 & 0.40 & \textbf{0.53} \\
Qwen3-4B-J& \textbf{0.53} & \underline{0.50} & \textbf{0.56} & \textbf{0.75} & 0.45 & \underline{0.45} & \underline{0.50} & \underline{0.50} & \textbf{0.57} & \underline{0.52} \\
\hdashline
Qwen3-8B& 0.45 & 0.45 & \underline{0.50} & \underline{0.46} & 0.46 & \textbf{0.55} & \textbf{0.55} & 0.20 & 0.42 & \textbf{0.50} \\
Qwen3-8B-PD& \textbf{0.57} & \underline{0.50} & \textbf{0.57} & \textbf{0.65} & \textbf{0.60} & \underline{0.45} & 0.40 & \textbf{0.90} & \underline{0.55} & \textbf{0.50} \\
Qwen3-8B-J& \underline{0.52} & \textbf{0.60} & 0.43 & 0.43 & \underline{0.48} & \underline{0.45} & \underline{0.50} & \underline{0.70} & \textbf{0.60} & \textbf{0.50} \\
\bottomrule
\end{tabular}
\captionof{table}{Detailed results on GameBench.}
\label{tab:gamebench_detail}
\end{minipage}
\end{center}

\clearpage

\section{Potential Applications of GARL}
\label{app:scenario_mapping}

\begin{center}
\begin{minipage}{\textwidth}
\centering
\scriptsize
\setlength{\tabcolsep}{2.2pt}
\renewcommand{\arraystretch}{1.15}
\begin{tabularx}{\textwidth}{
>{\raggedright\arraybackslash}p{0.115\textwidth}
>{\raggedright\arraybackslash}p{0.13\textwidth}
>{\raggedright\arraybackslash}p{0.14\textwidth}
>{\raggedright\arraybackslash}X
>{\raggedright\arraybackslash}p{0.105\textwidth}
>{\raggedright\arraybackslash}X
>{\raggedright\arraybackslash}p{0.105\textwidth}
}
\toprule
\textbf{Scenario} 
& \textbf{Candidate Set}
& \textbf{Allocating Agents}
& \textbf{Allocation Utility}
& \textbf{Arbiter}
& \textbf{Arbitration Utility}
& \textbf{Output} \\
\midrule
Issues-in-dispute ranking
& Candidate legal issues
& Prosecution and defence
& Strategic favourability, induced salience, and expected judicial recognition
& Judge
& Judicial adequacy and party-induced strategic emphasis
& Ranked issues in dispute \\
\midrule
Policy issue prioritisation
& Policy issues or reform proposals
& Stakeholder groups, departments, or advocacy agents
& Preference alignment, urgency, constituency benefit, and advocacy-induced salience
& Policy committee or decision authority
& Public value, feasibility, fairness, and stakeholder-induced salience
& Ranked policy priorities \\
\midrule
Investment allocation
& Investment targets, sectors, or projects
& Analyst agents with different strategies or risk preferences
& Expected return, risk exposure, portfolio fit, and strategic attention
& Investment committee or portfolio manager
& Return potential, risk control, diversification, and opportunity salience
& Ranked investment opportunities \\
\midrule
Risk assessment prioritisation
& Risks, incidents, or vulnerabilities
& Operational, legal, financial, or security risk agents
& Expected loss, likelihood, exposure, and domain-specific salience
& Risk committee or compliance authority
& Severity, controllability, regulatory importance, and cross-agent salience
& Ranked risk items \\
\midrule
Paper review prioritisation
& Submissions or review-relevant concerns
& Reviewer agents with different expertise or evaluation focuses
& Novelty, soundness, significance, clarity, reproducibility, and reviewer confidence
& Area chair or programme committee
& Venue criteria, review quality, confidence, disagreement, and reviewer-induced salience
& Ranked papers or review priorities \\
\bottomrule
\end{tabularx}
\captionof{table}{Illustrative mapping between GARL components and potential strategic prioritisation scenarios. Only issues-in-dispute ranking is empirically instantiated in this work; the remaining scenarios illustrate possible adaptations of the allocation--arbitration formulation.}
\label{tab:garl_scenario_mapping}
\end{minipage}
\end{center}

\clearpage

\section{Training Dynamics Figures}
\label{app:td}

\begin{center}
\includegraphics[width=\columnwidth]{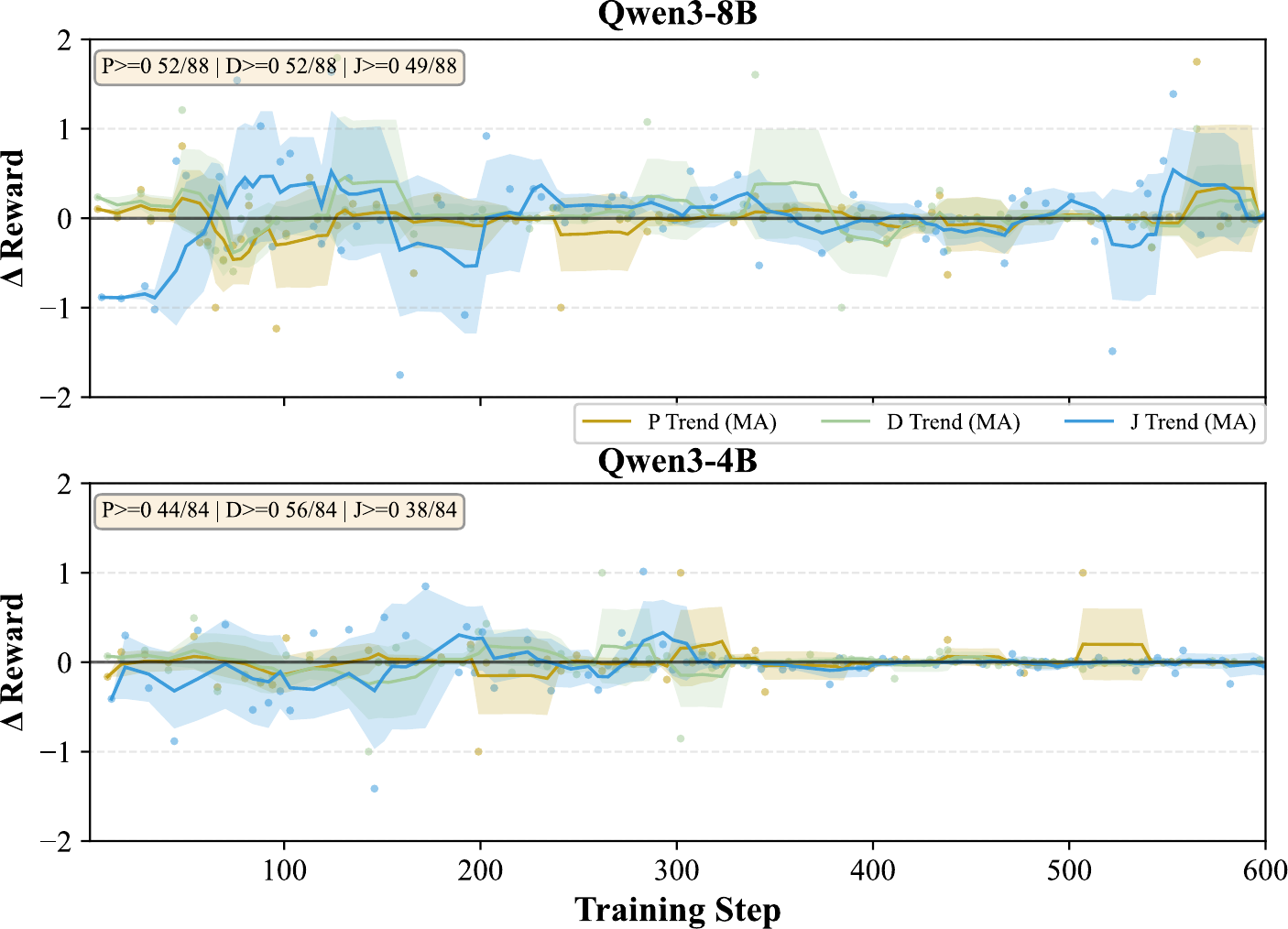}
\captionof{figure}{
Reward-change dynamics during GARL training.
The curves show moving-average trends of reward changes for the prosecution (\(P\)), defence (\(D\)), and judge (\(J\)) on Qwen3-8B and Qwen3-4B.
}
\label{figs/delta_reward}
\end{center}

\vspace{20pt}

\begin{center}
\includegraphics[width=\columnwidth]{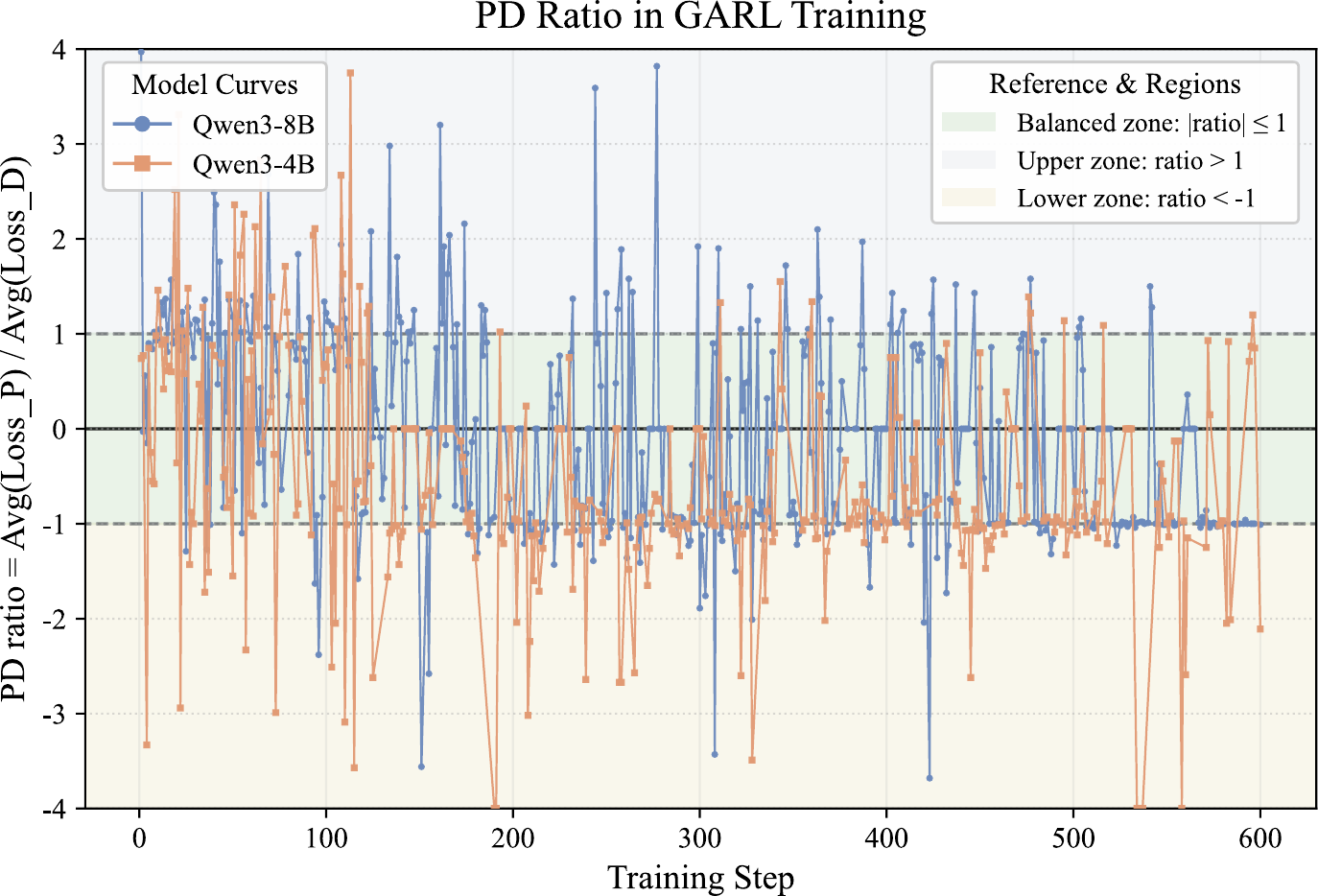}
\captionof{figure}{
Prosecution--defence loss-ratio dynamics during GARL training.
The ratio reflects both update direction and relative strength.
}
\label{figs/PD_ratio}
\end{center}

\clearpage

\section{Prompts}
\label{app:prompts}

\subsection{Utility and Data-Construction Prompts} 

\begin{center}
\includegraphics[width=\columnwidth]{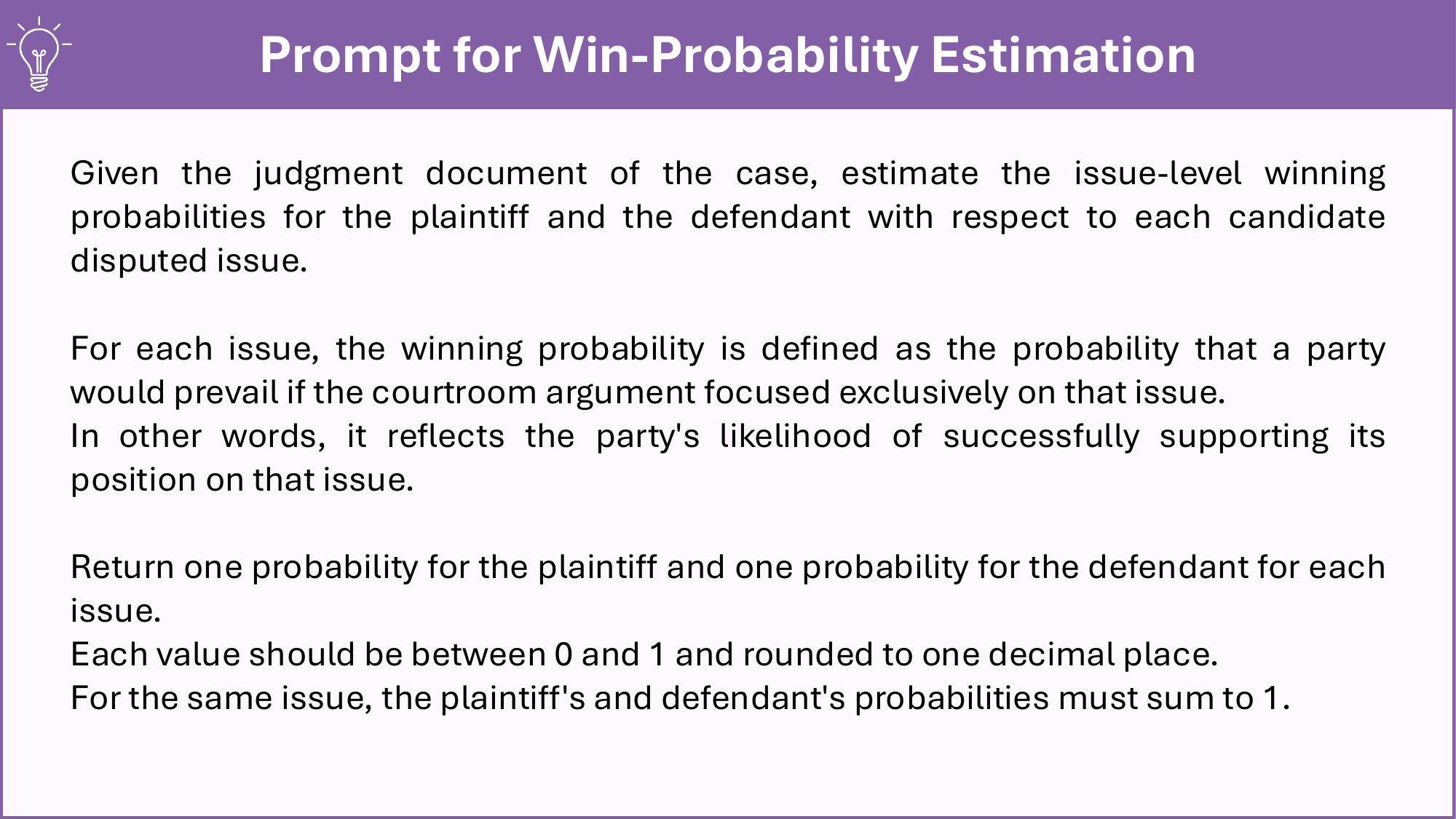}
\captionof{figure}{Prompt used for issue-level win-probability estimation.}
\label{figs/pm_win}
\end{center}

\begin{center}
\includegraphics[width=\columnwidth]{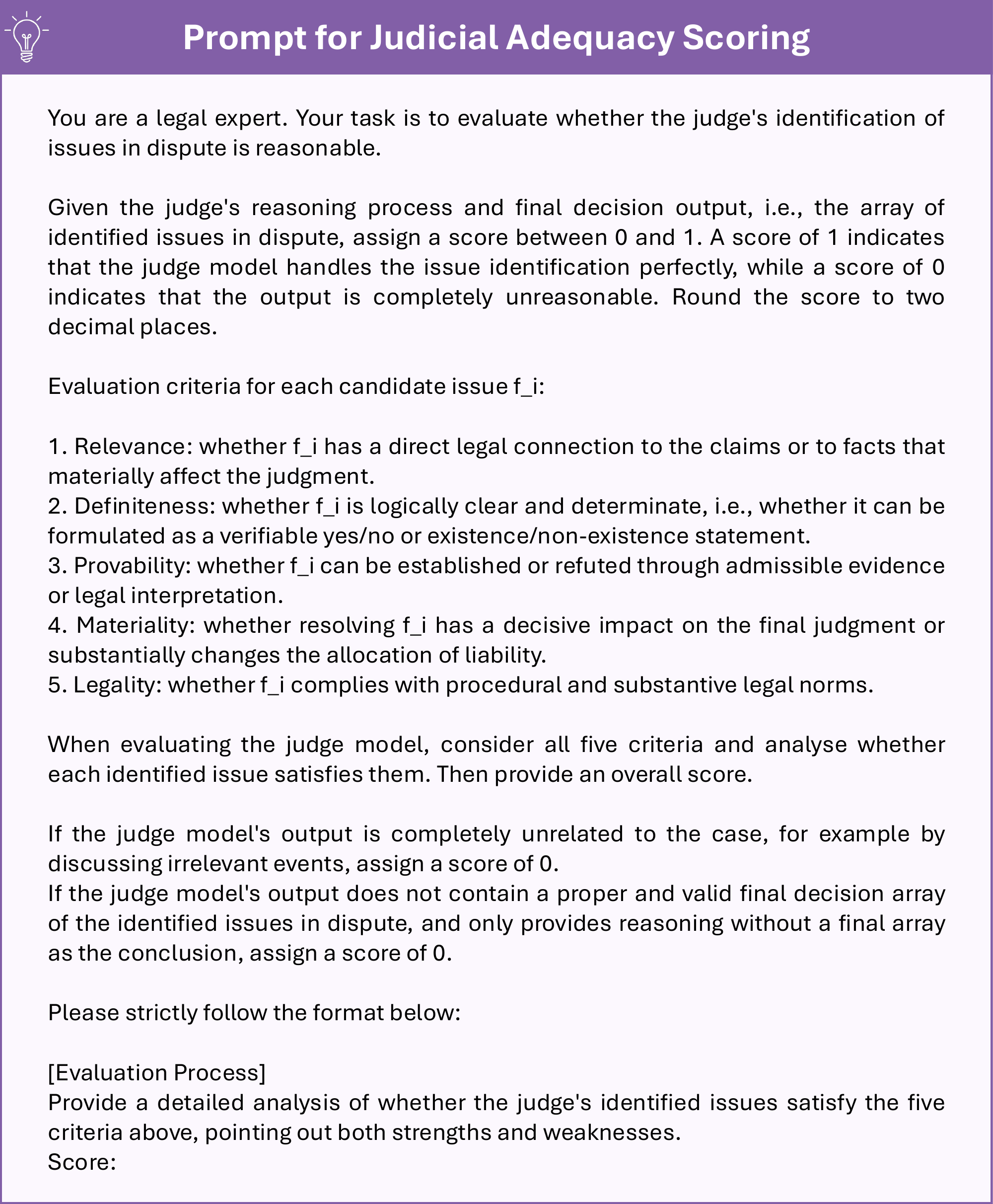}
\captionof{figure}{Prompt used for judicial adequacy scoring.}
\label{figs/pm_qi}
\end{center}

\begin{center}
\includegraphics[width=\columnwidth]{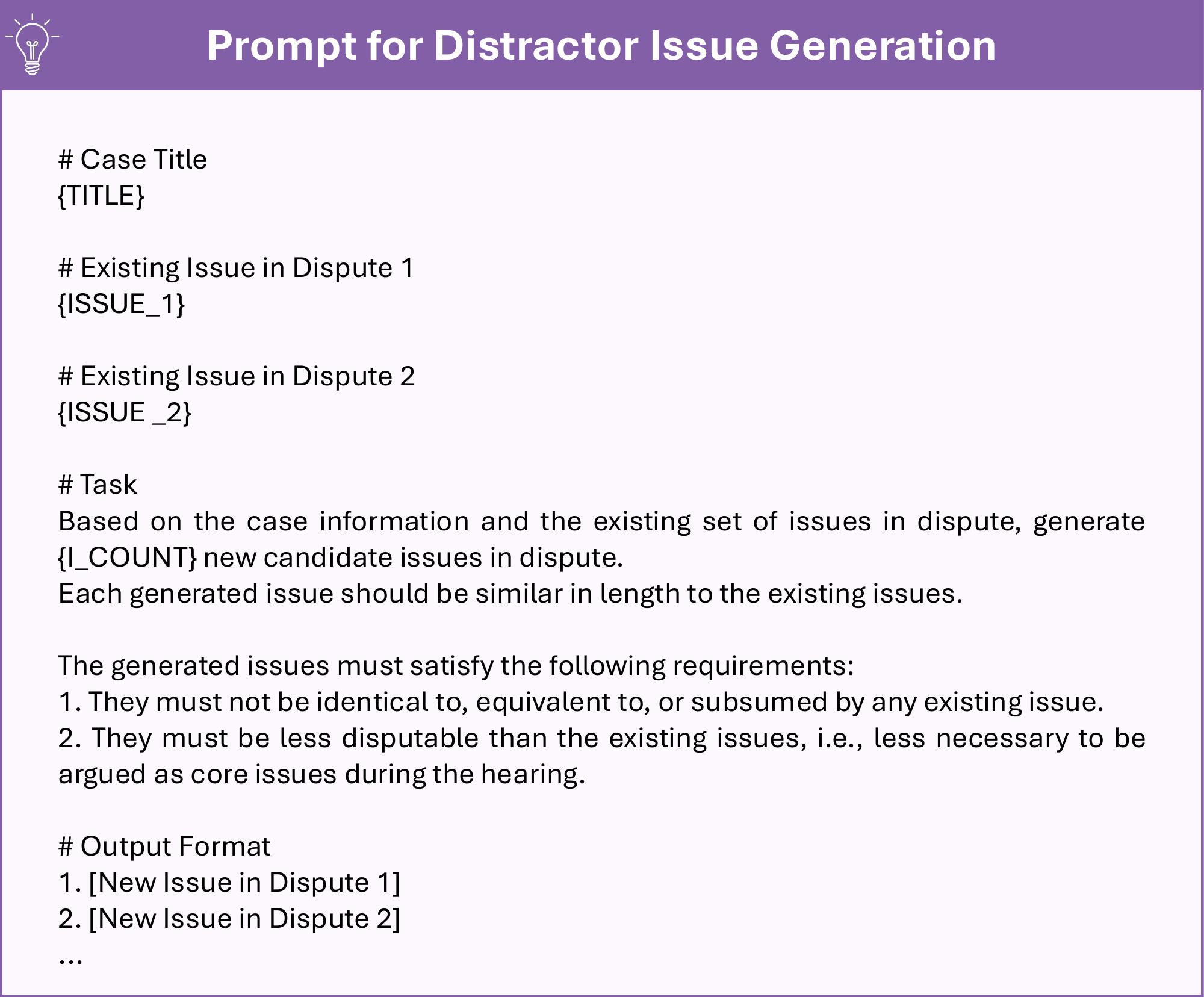}
\captionof{figure}{Prompt used for distractor issue generation.}
\label{figs/pm_gen}
\end{center}

\subsection{Training and Inference Prompts}

\begin{center}
\includegraphics[width=\columnwidth]{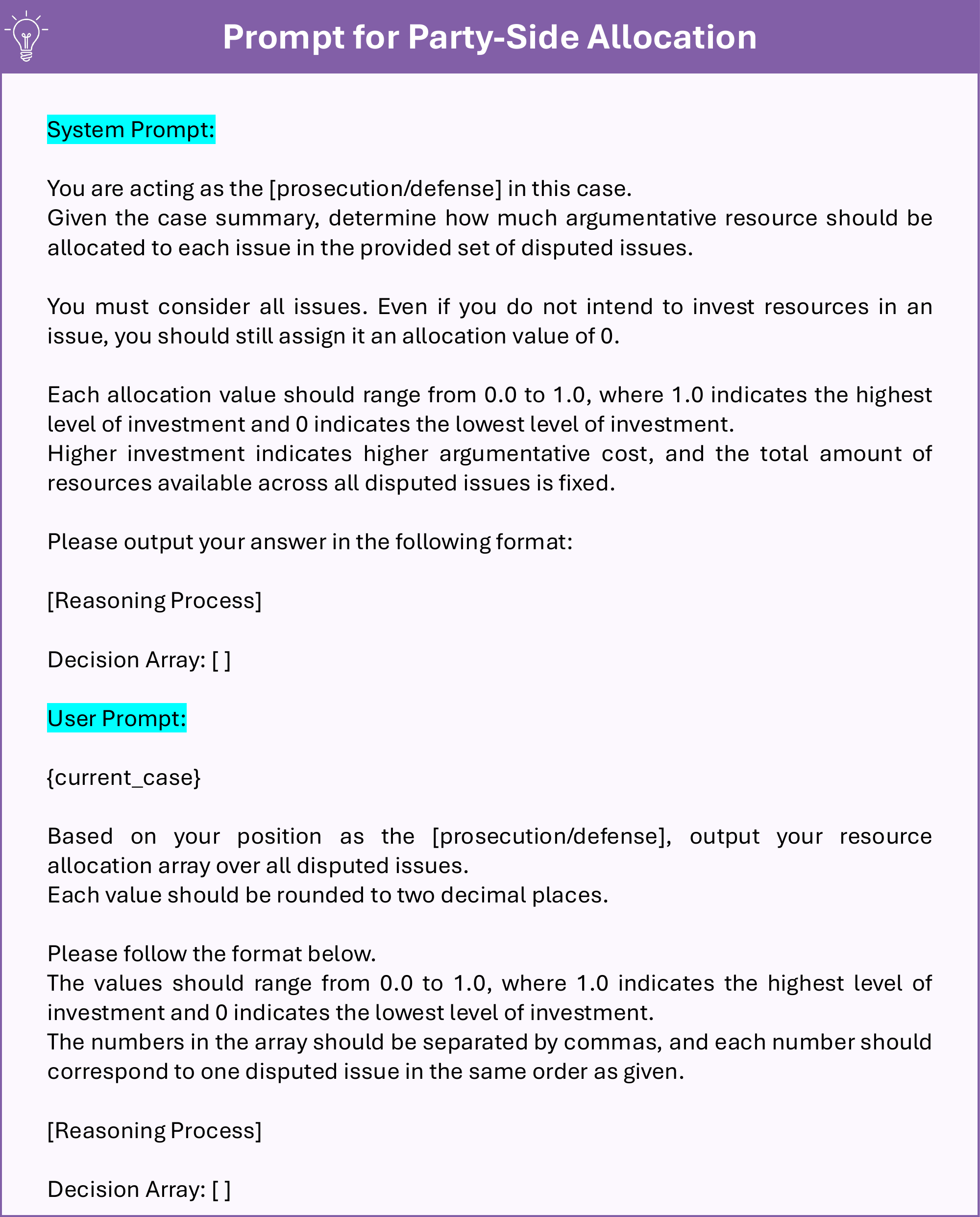}
\captionof{figure}{Prompt used for party-side allocation.}
\label{figs/pm_pa}
\end{center}

\begin{center}
\includegraphics[width=\columnwidth]{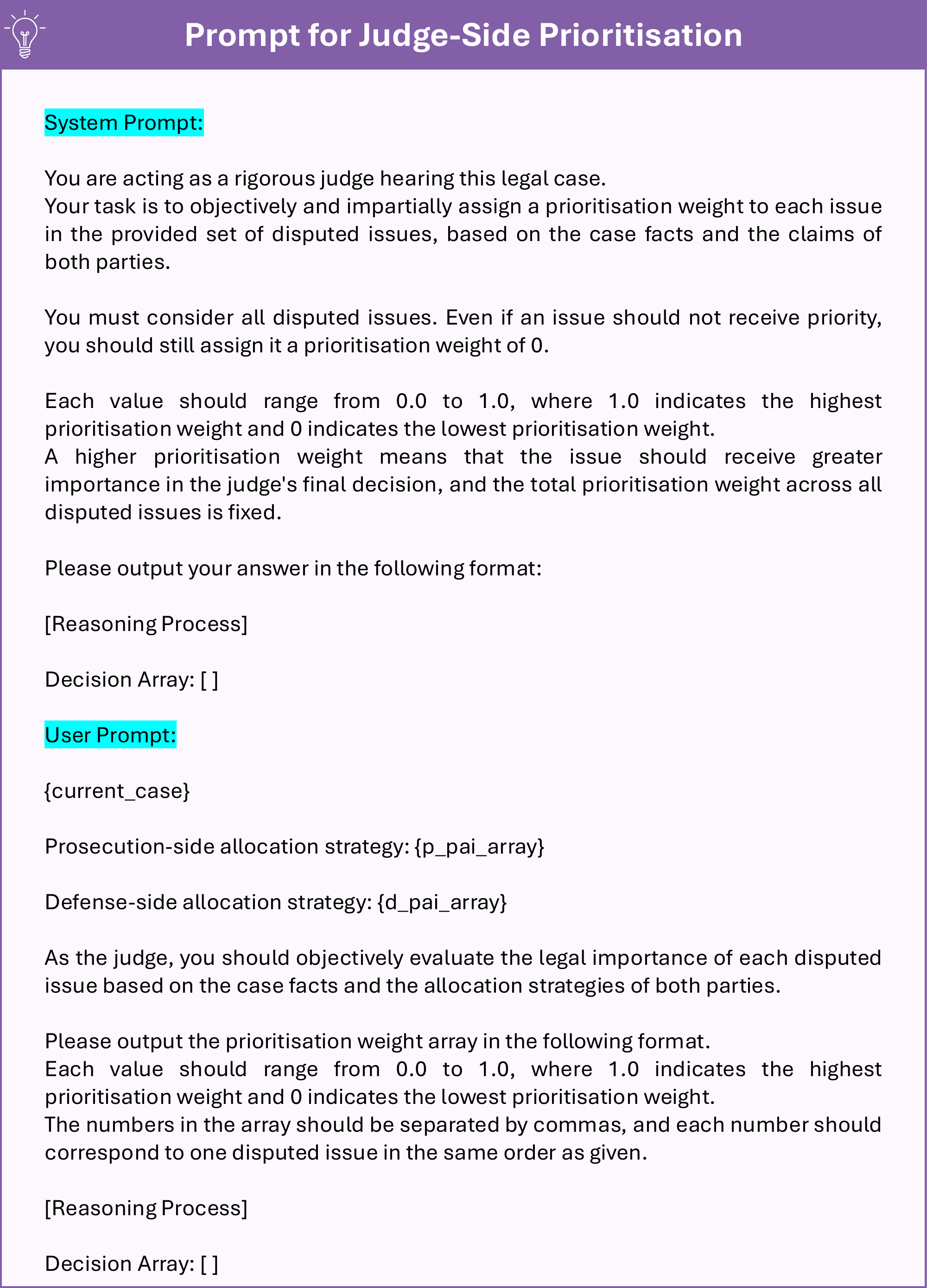}
\captionof{figure}{Prompt used for judge-side prioritisation.}
\label{figs/pm_j}
\end{center}

\begin{center}
\includegraphics[width=\columnwidth]{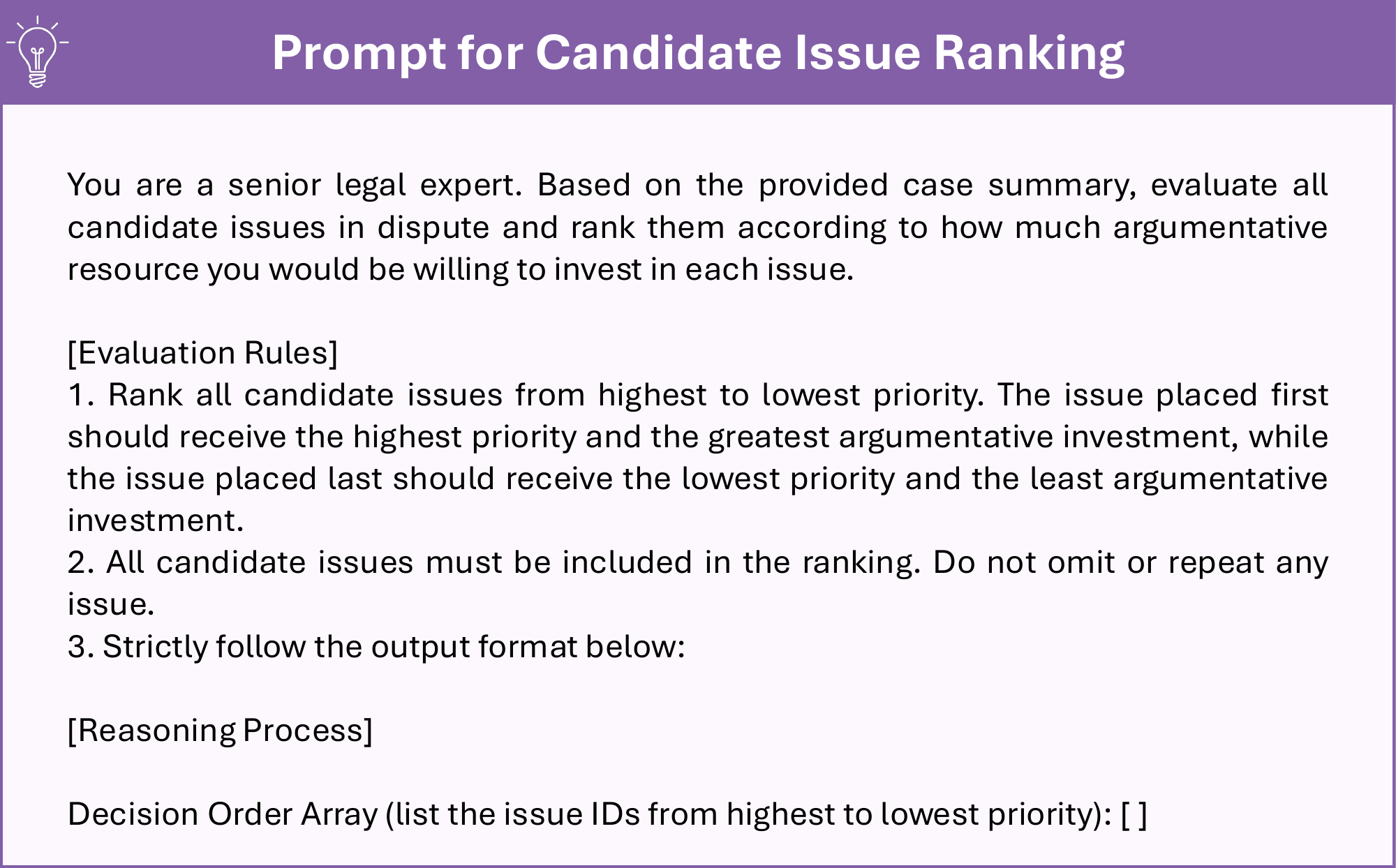}
\captionof{figure}{Prompt used for candidate issue ranking.}
\label{figs/pm_test}
\end{center}

\end{document}